%% file: RefineNet.tex
\documentclass{article}

\PassOptionsToPackage{numbers, compress}{natbib}
\usepackage[final]{arxiv}
\usepackage[10pt]{moresize}

\usepackage[utf8]{inputenc} 
\usepackage[T1]{fontenc}    

\usepackage{url}            
\usepackage{booktabs}       
\usepackage{amsfonts}       
\usepackage{nicefrac}       
\usepackage{microtype}      
\usepackage{lipsum}		
\usepackage{graphicx}

\usepackage{amssymb}

\usepackage{multicol}
\usepackage{amsmath}

\usepackage[colorlinks=true]{hyperref}
\usepackage{xcolor}

\usepackage[linewidth=1pt]{mdframed}

\usepackage{graphicx}
\usepackage{epstopdf}
\usepackage{subcaption}
\usepackage{multirow}
\usepackage{float}
\usepackage{empheq}

\usepackage{upgreek}
\usepackage{overpic}
\usepackage[]{multirow}
\usepackage[]{nicefrac}
\usepackage[]{bm}

\usepackage{wrapfig}

\usepackage{tikz}
\usepackage{pgfplots}
\definecolor{skyblue1}{rgb}{0.447,0.624,0.812}
\definecolor{scarletred1}{rgb}{0.937,0.161,0.161}
\definecolor{chameleon1}{rgb}{0.541,0.886,0.204}
\definecolor{chameleon2}{rgb}{0.441,0.586,0.204}

\floatstyle{boxed}
\newfloat{nomenclature}{thp}{lop}

\newcommand {\CoolNet}{{ContinuousNet}}
\newcommand {\CoolNets}{\CoolNet{}{s}}

\makeatletter
\DeclareRobustCommand\onedot{\futurelet\@let@token\@onedot}
\def\@onedot{\ifx\@let@token.\else.\null\fi\xspace}

\makeatother

\usepackage[verbose=true,letterpaper]{geometry}
\AtBeginDocument{
  \newgeometry{
    textheight=9in,
    textwidth=6.6in,
    top=1in,
    headheight=14pt,
    headsep=25pt,
    footskip=30pt
  }
}

\title{Continuous-in-Depth Neural Networks}

\date{} 					

\author{Alejandro F. Queiruga \thanks{Equal contributions.} \space  \thanks{Work also performed while previously at Lawrence Berkeley National Lab}	 \\ Google Research  \\ \textit{afq@google.com}
\And 
N. Benjamin Erichson$^*$ \\ ICSI and UC Berkeley \\ \textit{erichson@berkeley.edu}
\And 
Dane Taylor\\ University at Buffalo, SUNY \\ \textit{danet@buffalo.edu}
\And 
Michael W. Mahoney\\ ICSI and UC Berkeley \\ \textit{mmahoney@stat.berkeley.edu}}

\begin{document}

	\maketitle
	\begin{abstract}
		Recent work has attempted to interpret residual networks (ResNets) as one step of a forward Euler discretization of an ordinary differential equation, focusing mainly on syntactic algebraic similarities between the two systems. Discrete dynamical integrators of continuous dynamical systems, however, have a much richer structure.
		We first show that ResNets fail to be meaningful dynamical integrators in this richer sense.
		We then demonstrate that neural network models can learn to represent continuous dynamical systems, with this richer structure and properties, by embedding them into higher-order numerical integration schemes, such as the Runge Kutta schemes.
		Based on these insights, we introduce \CoolNet{} as a continuous-in-depth generalization of ResNet architectures.
		\CoolNets{} exhibit an invariance to the particular computational graph manifestation.
		That is, the continuous-in-depth model can be evaluated with different discrete time step sizes, which changes the number of layers, and different numerical integration schemes, which changes the graph connectivity. 
		We show that this can be used to develop an incremental-in-depth training scheme that improves model quality, while significantly decreasing training time.
		We also show that, once trained, the number of units in the computational graph can even be decreased, for faster inference with little-to-no accuracy drop. 
	\end{abstract}

\setcounter{footnote}{0}
\section{Introduction}

Among the many remarkable developments in machine learning (ML) that have taken place within the last decade, residual neural networks (ResNets)~\cite{he2016deep,he2016identity} and their variants~\cite{zagoruyko2016wide,huang2017densely,larsson2016fractalnet,gomez2017reversible,chang2018reversible,zhang2017polynet,yu2018learning} are especially prominent. 
They enable the training of very deep neural networks (NNs), and they have emerged as the state-of-the-art architecture for a large number of tasks in computer vision, natural language processing, and related areas.
Due to their popularity, there is a rich body of work that aims to explain ``why'' shortcut connections between layers enables the training of very deep NNs~\cite{veit2016residual,balduzzi2017shattered,hardt2016identity,li2017convergence,liu2019towards,li2018visualizing}.
Despite their advantages, ResNets also have various shortcomings, similar to other NN architectures. 
For example, they are known to be sensitive and brittle to hyperparameter choices during training, to various adversarial environments, and to other out-of-distribution changes during inference.
Moreover, any ResNet is generally difficult to compress without significantly decreasing model accuracy~\cite{wang2019haq,cheng2017survey,stock2020and}.

Recent work has established an interesting connection between ResNets and dynamical systems~\cite{haber2017stable,chang2017multi,zhang2019towards,ruthotto2019deep,zhang2019anodev2,zhang2019approximation}.
From this point of view, residual units are interpreted as one step of a forward Euler method for the numerical integration of a parameterized ordinary differential equation (ODE).
The ResNet residual unit  
of \cite{he2016identity},
\begin{equation}
\label{eq:resnet_unit}
x_{k+1} = x_k + \mathcal{R}(x_k; \theta_k),
\end{equation}
uses a ``skip'' connection where $\mathcal{R}$ is a map from the state $x_k$ to an update for $x_{k+1}$, where each unit $k$ is parameterized by weights $\theta_k$.
Eq.~\eqref{eq:resnet_unit} algebraically resembles a forward Euler integrator if $\mathcal{R}$ is a temporal rate of change,
\begin{equation}
\label{eqn:forward_euler_form}
x_{k+1}=x_k + \Delta t \,\, \mathcal{R}(x_k;\theta_k),
\end{equation}
with a time step size $\Delta t=1$. Here, $\theta_k=\theta(t_k)$ denotes the parameters at time $t_k\equiv k\Delta t$, where the unit index now corresponds to a time step index, $k=0,1,\dots,N_t$.
In this analogy, each $x_k\approx x(t_k)$ is an approximation to a continuous trajectory $x(t)$ at time $t_k$. 
The final output at $t=T$ of the ResNet is $x(T)$ for $\Delta t=T/N_t$. 
Clarifying this analogy appropriately is one of the main contributions of this work.

Going beyond the observation of simply an algebraic resemblance, in this paper, we ask three main questions: 
\begin{itemize}
\item
How exactly does a residual unit in a ResNet correspond to a single step of the forward Euler method?  
\item
Is there a more appropriate interpretation for this class of NNs that stems from dynamical systems and numerical integration theory?  
\item
If so, can we use this improved understanding to develop improved architectures, e.g., for improved training and/or inference?
\end{itemize}
While many recent developments have used a perceived connection between ResNets and dynamical systems as inspiration---most notably ODE-based models such as Neural ODEs~\cite{chen2018neural}---thus far there has been a surprising lack of empirical or theoretical work that fully investigates these questions. 

To address these questions, we build upon the dynamical systems view of ResNets by laying out the properties (which are well-known in the numerical integrator community) that a numerical integrator must satisfy.
With these properties, we see that algebraic similarity is \emph{not} sufficient for a model to be a numerical integrator in a meaningful sense. 
Importantly, the learning process itself is a confounding factor: what matters most is how the model behaves after it is trained, not its structural form before it sees any training data.
Our main results answer our three questions:
\begin{itemize}
\item
We show that, ResNets (and ODE models based on forward Euler) fall short of having the complete properties of being numerical integrators. In other words, although they are good discrete models, ResNets fall short of being interpretable as numerical approximations to continuous models.
Thus, the connection between them and continuous dynamical systems is superficial at best.
\item
We show these properties are satisfied by models where the NN is embedded inside of a computational graph that is defined by a higher-order integrator (e.g., a Runge-Kutta method).
Informally, this occurs because higher-order integrators introduce an inductive bias, which leads to a NN that more-strongly inherits the smoothness property of a continuous-time dynamics.

\item
We show that these insights can be used to develop \CoolNet{}, a continuous-in-depth generalization of ResNets.
\CoolNets{} have improved robustness properties due to being meaningful approximations to continuous models, as opposed to being just stand-alone discrete models.
In particular, these robustness properties can be exploited to develop improved methods for both training and inference.
\end{itemize}

\paragraph{Discrete NN Models Correspond to Fixed Discrete Computational Graphs.}

ResNets and other traditional discrete NNs can be described by a discrete computational graph that represents the stages of computation and data flow during a forward evaluation.
In this view, the basic ResNet unit corresponds to one piece of the computational graph of the whole model.
Each node in the computational graph corresponds to an operation with an associated set of trainable weights, and each edge corresponds to some intermediate state.
Importantly, in this setup, nodes represent {\em both} computation and parameters.
Hence, once a model is trained, there is no prescribed way to significantly alter the computational graph.

Of course, specific architectures (i.e., specific designs of NNs for a given problem) often correspond to a family of graphs with some hyperparameters that generate a particular computational graph.
For example, the ResNet architecture uses the total number layers as a hyperparameter.
The choice of this hyperparameter dictates how many times to repeat the basic residual unit, which in turn generates one instance of a NN computational graph whose weights will then be trained.
From this perspective, we say that for any given hyperparameter choice, an architecture \emph{manifests} a computational graph.
That is, with depth=100, a ResNet architecture manifests a graph known as ResNet-100.
However, because parameters and computations are tied together, each of these graph manifestations is really best thought of as a different discrete model.
For instance, a ResNet-100 cannot be easily transformed into a ResNet-50 or a ResNet-200, or even a ResNet-101.

\begin{wrapfigure}{R}{0.50\textwidth}\vspace{-0.3cm}
	\centering
	\includegraphics[width=0.9\linewidth]{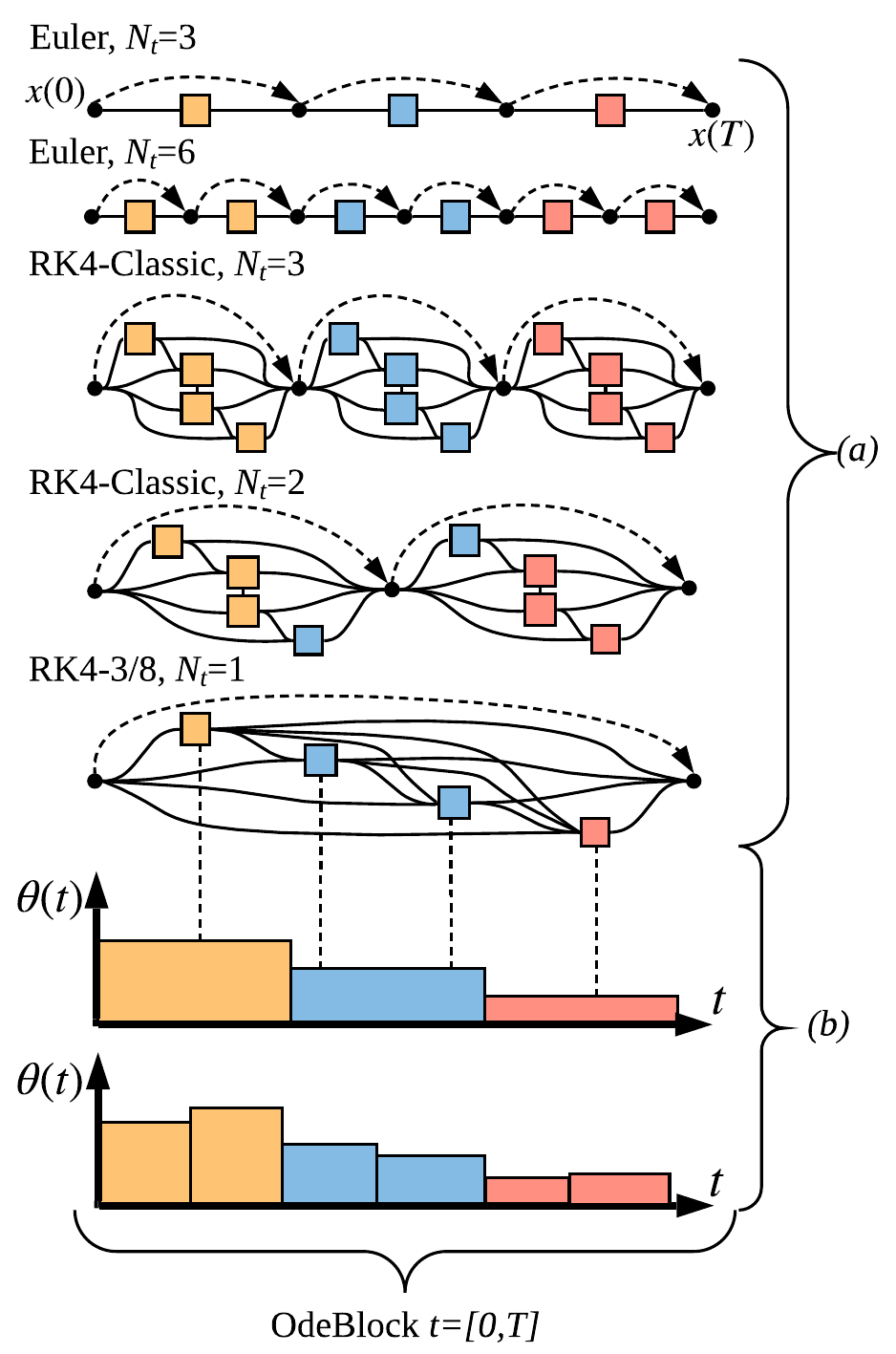}
	\caption{\CoolNets{} are \emph{continuous-in-depth} models that are trained to obey the properties of numerical integrators (i.e., discrete approximations to continuous operators).
		They use decoupled representations of the weights $\theta(t)$ and the {computational graph} (i.e., the number and connectivity of NN layers) to allow \emph{manifestation invariance}:
		(a) Different discrete graphs can be freely interchanged to implement the integration for a given weight function
		$\theta(t)$;
		(b)	$\theta(t)$ is itself represented by trainable parameters and can be refined (or compressed) independently of the computation graph.
		The simplest OdeBlock corresponds to a ResNet block (top row).
	}\vspace{-0.4cm}
	\label{fig:\CoolNet_graph}
\end{wrapfigure}

\paragraph{Discrete Weights versus Parameterized Weight Functions.}

For our purposes, we choose to parameterize the ODE corresponding to Eq.~\eqref{eqn:forward_euler_form} as
\begin{equation}\label{eq:one}
\frac{\mathrm{d} x(t)}{\mathrm{d} t} = \mathcal{R}(x(t),\theta(t;\hat{\theta}))  .
\end{equation}
This defines the continuous evolution from an initial condition $x(0) = x_{in}$ to the final output $x(T)=x_{out}$.
Here, $\theta(t;\hat{\theta})$ denotes continuously-varying weight functions that parameterize the ODE and that themselves are parameterized by a countable tensor of trainable parameters $\hat{\theta}$.
The parameter $t\in[0,T]$ in this ODE represents time as an analogy to the depth of the NN. Using a finer temporal discretization, with a smaller $\Delta t$, corresponds to deeper network architectures.
We highlight a subtle but important difference between Eq.~\eqref{eq:one} and other works in the literature (such as those discussed in Sec.~\ref{sxn:related-work}).
Previous works define the ODE using $\mathcal{R} \equiv \mathcal{R}(x(t),t;\hat{\theta})$, which does not have a clear description of parameters-in-time.
In contrast, our formulation in Eq.~\eqref{eq:one} makes explicit how the update map $\mathcal{R}$ varies with time $t$, in particular through a set of parameterized time-varying functions $\theta(t;\hat{\theta})$.
%
This formulation allows us to make a clearer connection to ResNets, for which $x(0)$, $x(T)$, and $x(t_k)$ are interpreted as the input, output, and intermediate states of a ResNet block.

We consider $\mathcal{R}(x,\theta)$ as a pure mathematical function which takes parameters as inputs, as opposed to a subgraph of a discrete NN that ``owns'' parameters.
Its input $x$ accepts anything that evaluates as a correctly shaped tensor, and its input $\theta$ must evaluate as a collection-of-tensors that matches the sub-elements of $\mathcal{R}$ (e.g., the convolutional kernel weights). 
For a discrete ResNet model, the inputs are: a tensor of activations $x=x_k$ in layer $k$; and a collection-of-tensors of trainable parameters $\theta=\theta_k$ that are unique to the layer $k$. 
For a continuous model, the inputs are instead continuous functions $x=x(t)$ and $\theta=\theta(t)$, which can be evaluated at any real-valued $t\in[0,T]$. 
To seek an approximate model, i.e., a discrete model that meaningfully approximates a continuous model, the weight functions $\theta(t) = \theta(t;\hat{\theta})$ can themselves be parameterized by trainable parameters $\hat{\theta}$.
In general, these parameters $\hat{\theta}$ can be used differently, and can have different sizes/shapes, than the ResNet's weights $\theta_k$. 
In summary, for a continuous model, we do not train a discrete set of weights, but rather, we train discrete parameterizations of continuous weight functions.

\paragraph{\CoolNets{} are Continuous-in-Depth NN Models.}

Based on these insights, we propose \CoolNets{} as a broad family of NN models that use a \emph{continuous-in-depth} residual block, henceforth called an ``OdeBlock.''
OdeBlocks provide a generalization of ResNet blocks, and they can operationally serve as a 1-to-1 replacement for them.
Crucially, OdeBlocks are designed to learn a continuous model, which in turn decouples the representations for weights and the computational graph. 
Performing a forward pass of a \CoolNet{} involves numerical integration of Eq.~\eqref{eq:one} for a prescribed function $\theta(t)$.
We represent the continuous function $\theta(t)$ using a set of basis functions $\phi^\beta(t)$ with trainable coefficients $\hat{\theta}^\beta$, with $\beta = 0,1,\dots, M$. 
In the simplest setting, we can choose the forward Euler integrator and let $\phi^\beta(t)$ be piecewise-constant functions, in which case, an OdeBlock recovers a ResNet block.
More generally, choosing higher-order numerical integrators and/or other basis functions enables great flexibility to separate the computational graph from the choice of parameters, thereby permitting us to develop models that have a meaningfully continuous depth.

\paragraph{\CoolNets{} Can Be Represented by Many Discrete NN Models.}

An important consequence of ensuring that a discrete NN model approximates a continuous dynamical system is that it is then possible to specify more than just one series of discrete computational graphs to evaluate the continuous model. 
That is, a given \CoolNet{} model can be evaluated---with no or very little loss in quality---with many different discrete computational graphs, i.e., with many different discrete NN models. (By analogy, the integral of an ODE can be approximated by many different numerical integrators,  each of which can be implemented with different time-step sizes.)
We will exploit this flexibility for improved training as well as for improved inference.

The following are examples of how different computational graph manifestations can arise for \CoolNets{}:
we can choose different numerical integrator schemes (e.g., forward Euler versus Runge-Kutta, leading to \emph{integrator refinement});
we can choose different time steps ($\Delta t$, which correspond to different depth in the discrete model, leading to \emph{time-step/depth refinement}); and
we can choose different basis function sets to approximate the weight function $\theta(t;\hat{\theta})$ (leading to \emph{weight refinement}).
In each case, we obtain different discrete sets of parameters that represent the same continuous model.
As an illustration, Fig.~\ref{fig:\CoolNet_graph} illustrates how a whole family of equivalent deep NN graphs based on numerical integrators can be generated for one \CoolNet{} model, as well as how the parameters of the model can be projected onto different computational graphs. 

We call this property \emph{manifestation invariance}: namely, that a trained \CoolNet{} model can be evaluated through many different routes; and, given a set of input features for a \CoolNet{} model, there are many distinct series of discrete computations that produce approximately the same outputs.
More generally, manifestation invariance refers to the ability to specify systematically many non-trivially distinct computational graphs that evaluate the same continuous model.%
\footnote{Examples of trivial alterations to a NN are adding identity layers/activations in their linear regime or sandwiching a matrix and its inverse next to each other. 
To be meaningfully manifestation invariant, we refer to being able to ``rewrite'' a section of the computational graph and parameters into another form, e.g., with integrator refinement or time-step/depth refinement or weight refinement.} 
Importantly for applications, only the trained instance of the model needs to be given, i.e., the transformations can be performed without data.
Given one graph, both ``shrinking'' (manifesting a smaller graph with fewer parameters) and ``growing'' (manifesting a larger graph with more parameters) operations are both possible.%
\footnote{It is also possible to manifest a graph with more computations but with fewer parameters, or vice versa.}

\begin{figure}[!t]
    \centering
\fbox{%
    \parbox{\textwidth}{%
 \begin{multicols}{2}
 \begin{description}
 \item[$k$] An index into depth or a discrete series of steps.
 \item[$t$] Time, or continuous depth.
 \item[$T$] Maximum time.
 \item[$t_k$] Time at index $k$.
 \item[$\Delta t$] Time step size, $t_{k+1}-t_k$.
 \item[$x$] Signal as a free variable.
 \item[$x_k$] Signal at step index $k$.
 \item[$x(t)$] Signal as a function of time.
 \item[$\theta$] Weights as a free variable.
 \item[$\hat{\theta}$] Tensor of discrete trainable parameters.
 \item[$\hat{\theta}_k$] Weights at step index $k$.
\item[$\theta(t)$] Continuous weight function in time.
 \item[$\theta(t;\hat{\theta})$] Continuous weight function in time explicitly parameterized by $\hat{\theta}$.
 \item[$\phi^\beta(t)$] Basis function of coefficient $\beta$.
 \item[$\hat{\theta}^\beta$] Coefficient of $\theta(t;\hat{\theta})$ corresponding to $\phi^\beta$.
 \item[$N_t$] Number of time steps.
 \item[$M$] Number of basis functions.
 \item[$\mathcal{R}(x,\theta)$] Residual module evaluated with input signal $x$ and parameters $\theta$.
 \item[$F(\cdot;\hat{\theta})$] An arbitrary NN with trainable parameters.
 \item[$G(\cdot;\hat{\theta})$] The specific shallow $tanh$ NN used to learn pendulum dynamics in Sec.~\ref{sec:overview}.
 \item[$f$] Known functional form of $\dot{x}$ that could be used for numerical integration.
 \item[ {$func[a](x)$} ]  Partial function application to $a$ and then $x$.
 \end{description}
 \end{multicols}
     }%
}
 \caption{Nomenclature.}\label{fig:Nomenclature}
\end{figure}

In summary, the key innovations of \CoolNets{} are as follows:
	
\begin{itemize}
\item 
An ODE formulation, combined with a basis function representation of weight functions,  \emph{decouples the parameters of a model from its computational graph.}
\item 
At training, the number of layers and trainable parameters can be increased/decreased for a given \CoolNet{}, thereby enabling stable incremental deepening during training.
\item 
At inference, a trained \CoolNet{} can be re-manifested as a different discrete computational graph, and thus it can be compressed into a shorter/smaller network of similar accuracy without revisiting data.
\end{itemize}

The property of manifestation invariance allows \CoolNets{} to be adapted flexibly to different architectures during training, testing, and deployment. 
On the theoretical side, time-step refinements and integrator refinements can be guided using integration theory, and weight refinements can be guided using functional analysis. 
On the practical side, \CoolNets{} achieve state-of-the-art accuracy on standard computer vision problems, and these refinements lead to much more robust models at both training and inference, with little or no loss in model quality.

Research code and examples are available on GitHub: \href{https://github.com/afqueiruga/ContinuousNet}{https://github.com/afqueiruga/ContinuousNet}.

\paragraph{Outline of the Paper.}

The nomenclature in Fig.~\ref{fig:Nomenclature} defines the notation that is used throughout this manuscript.
After a brief review of related works in Sec.~\ref{sxn:related-work}, we will in Sec.~\ref{sec:overview} address the question of what it means for a discrete ResNet model to correspond to a discrete numerical approximation of a continuous ODE dynamical system.
Based on these insights, in Sec.~\ref{sxn:continuous-in-depth} we will introduce our \CoolNet{} model.
In Sec.~\ref{sxn:refinenet-empirical}, we will present empirical results illustrating the performance of the \CoolNet{} model, including how we can exploit the model's manifestation invariance property to perform weight refinement during the training process, and how we can vary the depth of the model and/or the specific numerical integrator during the inference step.
In Sec.~\ref{sxn:conc}, we will conclude with a brief discussion.

\section{Related Work}
\label{sxn:related-work}
	
The theory of dynamical systems and control provides powerful tools for studying and describing the qualitative behavior of linear and nonlinear systems that are expressed as differential equations. 
These tools are also useful for analyzing ML and optimization algorithms~\cite{weinan2017proposal,su2016differential,muehlebach2019dynamical,orvieto2019shadowing}, as well as for improving our understanding of NNs~\cite{howse1996gradient,lu2017beyond,li2017maximum,li2018optimal,benning2019deep,liu2019deep,sonoda2019transport}.
Along these lines, several physics-based models~\cite{ruthotto2019deep,long2018pde,rackauckas2020universal,lu2019deeponet,erichson2019physics,azencot2020forecasting} and continuous analogues to NNs~\cite{greydanus2019hamiltonian,cranmer2020lagrangian,bai2019deep} have recently been proposed. 
Further, the stability of the dynamics of ResNets has been explored~\cite{zhang2019forward,yang2017mean}, and this sheds light on the relationship between the depth of the architecture, regularization, and the stability of the feature space.

\paragraph{Dynamical Systems Perspective on ResNets.}
Related to our work is the interpretation of residual units as forward Euler steps of a dynamical system.
This connection was first established by~\cite{weinan2017proposal,haber2017stable}, and it has since been used as inspiration for the design of ResNet variants and algorithms, such as invertible networks~\cite{behrmann2019invertible} and stabilization \cite{haber2017stable,haber2019imexnet,ciccone2018nais}. 
Motivated by this perspective, \cite{chang2017multi} proposed a multi-level refinement scheme that greatly accelerates training by allowing new layers to be inserted into a ResNet that is derived from the forward Euler relation.
Lu et al.~\cite{lu2017beyond} extended the dynamical systems perspective on ResNets to a broader range of network architectures, such as PolyNet~\cite{zhang2017polynet}, FractalNet~\cite{larsson2016fractalnet} and RevNet~\cite{gomez2017reversible,chang2018reversible}. 
Specifically, they related these networks to other more sophisticated numerical integration schemes.
Through this lens, the $\epsilon$-ResNets~\cite{yu2018learning} have similarities to the midpoint method, and DenseNets~\cite{huang2017densely} have similar connectivity to the explicit Runge Kutta (RK4-3/8) method.  
Others have rediscovered integration schemes such as the implicit trapezoidal rule, which coincides with the scaled Cayley transform in the context of recurrent neural networks (RNNs)~\cite{helfrich2018orthogonal}.

\paragraph{Neural Ordinary Differential Equations.}
In recent literature, so-called Neural ODEs have attracted attention in both the ML and the physical sciences communities~\cite{chen2018neural,ruthotto2019deep,dupont2019augmented,zhong2019symplectic,zhang2019anodev2,gholami2019anode}.
The idea of Neural ODEs is to use NNs to parameterize a differential equation that governs the hidden states $x(t)\in \mathbb{R}^d$ with respect to time $t$,
\begin{equation}\label{eq:neuralODE}
	\dot{x} = F(x(t),t;\hat{\theta}) = \sigma(\hat{W} \dots \sigma(\hat{W}x+\hat{A}t+\hat{b}) \dots + \hat{b}),
\end{equation}
where $F: \mathbb{R}^d \rightarrow \mathbb{R}^d$ denotes the network that is parameterized by the learnable parameters  $\hat{\theta}=\{\hat{A},\hat{W},\hat{b},\dots \}$, and where $\dot{x}=d x/d t$ is the time derivative, and $\sigma$ is any nonlinearity.
In other words, Neural ODEs provide a framework to express residual-like networks by repeatedly applying a single unit, which can vary over time since it takes $t$ as an input.
The ``depth'' of the computation is the integration process from the input signal $x_0= x_{in}$ to the output $x(T) = x_{out}$:
\begin{equation}\label{eq:neuralODE2}
	x_{out} = x_{in} + \int_{0}^{T} F(x(t),t;\hat{\theta}) \mathrm{d}t,
\end{equation}
This equation is then solved with and ODE integrator 
\begin{align}
	x_{out} & \approx \text{AdaptiveODESolve}\left[F(x,t;\hat{\theta}),\, x_{in},\, t_{start}=0, t_{max}=T, tolerance=\epsilon \right].
\end{align}
In recent work, higher order adaptive methods, which automatically change the time step by estimating truncation error, have been developed~\cite{chen2018neural,zhuang2020adaptive,zhang2019anodev2}. 
In this work, we instead focus on fixed-time-step solvers to replicate ResNet, and we directly investigate properties of learning dynamical systems.

The properties of Neural ODEs are an active area of investigation~\cite{zhang2019approximation,massaroli2020dissecting,finlay2020train,hanshu2019robustness}, and a range of methods for training have been proposed~\cite{chen2018neural,quaglino2019snode,zhuang2020adaptive}.
Of particular interest is  adjoint-based back-propagation, which is very memory efficient but is impractical in practice because it requires more time steps and higher accuracy.
The proposed checkpoint method by~\cite{zhuang2020adaptive} mitigates some of these practical difficulties.
However, for most problems, auto-differentiation~\cite{paszke2017automatic} remains the best off-the-shelf approach for learning the parameters $\theta$.
Other theoretical research fronts include interfacing Neural ODEs with normalizing flows~\cite{grathwohl2018ffjord,yang2019pointflow,kohler2019equivariant,massaroli2020stable}, graph NNs~\cite{poli2019graph,sanchez2019hamiltonian}, stochastic differential equations~\cite{liu2019neural,jia2019neural,guler2019towards,li2020scalable} and RNNs~\cite{chang2018antisymmetricrnn,NIPS2019_8773,habiba2020neural,erichson2020lipschitz,lim2020understanding}.

One shortcoming of NeuralODEs~\cite{chen2018neural} and many of its extensions is that it does not specify a way to break the model down into a series of smaller building blocks.
Therefore, its internals are opaque to interpretation, and one cannot easily determine which parameters and operations correspond to the different stages of computation. 
That is, the NeuralODE framework evaluates  Eq.~\eqref{eq:neuralODE} using an ordinary network that is evaluated at different values of $t$.
The PDE-motivated models of~\cite{ruthotto2019deep} overcome this issue by assigning new parameters to individual time steps. However, that family of PDE-motivated models are essentially discrete ResNets, albeit with new types of residual units motivated by single-stage numerical integrators.

\section{Revisiting the Forward Euler Interpretations of Residual Networks}
\label{sec:overview}

Before we develop the foundations for continuous-in-depth NNs (in Sec.~\ref{sxn:continuous-in-depth}), in this section, we are going to address the question of what it means for a discrete model to approximate a continuous model, and in particular,  what it means for a discrete ResNet model to correspond to a discrete numerical approximation of a continuous dynamical system.
To do so, we revisit the concept of ODE-based neural networks (ODE-Nets) by examining how the choice of numerical integrator affects their performance. 
We study the application of different ODE-Net models to the task of learning to approximate a prototypical continuous-in-time ODE dynamical system.%
\footnote{Our use of the term ``ODE-Net'' is very general: an ODE-Net can be any NN with an ODE-like structure. The method described here is to train a NN to approximate one step of a discrete time series using a fixed-step integrator with traditional back-propagation to calculate gradients. This method is as opposed to optimizing over longer-trajectories using the adjoint method, which has been the set up in recent literature such as~\cite{chen2018neural}.}
A single-step ODE-Net produced with forward Euler corresponds to a ResNet unit.
Adopting this approach allows us to study whether a ResNet is really analogous in some way to the forward Euler integrator (or to some other discrete numerical integrator). 
If it is, then a ResNet should behave like that integrator for the task of approximating the ODE. 
If it is not, then it will not behave like a numerical integrator.
The generalization to ODE-Nets then determines the conditions needed to learn a model that does behave like a numerical integrator.

\subsection{Overview of General Approach}

To evaluate the dynamics properties of ResNets, as well as other models that are motivated by, or that purport, to represent continuous-in-time ODEs, we use the framework of ODE-Nets to study the following two questions:
\begin{itemize}
	\item 
	\textbf{Pre-training behavior:}
	How are the properties of a pre-trained model affected by time step changes in the~data?
	
	\item 
	\textbf{Post-training behavior:}
	How are the properties of a model affected by exchanging the numerical integration scheme during the inference step? 
\end{itemize} 

Both of these questions go well beyond looking at just training/testing curves for a single discrete NN model, as is typical in ML.
Instead, these questions require one to study consistency, convergence, and convergence rate properties (e.g., as the time step $\Delta t$, or depth $N_t$, is varied) in a parameterized sequence of (informally, discrete-to-continuous) models.

For example, one can look at the ``discretization error'' associated with the model (e.g., $E_k(\Delta t)$ or $E(\Delta t) $, described below and in Sec.~\ref{sec:overview-integrators}), as is common in scientific computing and the numerical analysis of numerical integrators.
In this case, it is well-known~\cite{leveque2007finite} that numerical integrators should have the following properties, as the time step $\Delta t$ (corresponding, for us, to the depth $N_t$) is varied:
\begin{itemize}
	\item
	{\bf Consistency:} 
	As the time step decreases, i.e., as $\Delta t\to0$, the ratio $E_k(\Delta t) / \Delta t$ decays to zero, where $E_k(\Delta t)=||x_k-x_{true}(k\Delta t)||$ is the local error at time $k\Delta t $.
	
	\item
	{\bf Convergence:} As $\Delta t \to 0$, the global error $E(\Delta t) = ||x_k-x_{true}(k\Delta t)||$ decays to zero.
	
	\item
	{\bf Convergence Rate:}  
	The global error $E(\Delta t) $ decays to zero as $E(\Delta t)=  \mathcal{O}(\Delta t^r)$, which depends on the integrator's convergence rate  $r$ (i.e., order of accuracy). The order of accuracy of the forward Euler method is $r=1$, while the midpoint method and RK4-classic are of order $r=2$ and $r=4$, respectively.
\end{itemize}

Hence, in order to be a numerical integrator in a meaningful sense, a model must satisfy these three properties, as the time step $\Delta t$ is decreased.
Simply having a similar syntactic algebraic similarity is not enough.
Most commonly, these properties require an implicit assumption that the numerical integrator is applied to a dynamical system that is continuous with respect to time.
In scientific computing, the derivative function $f$ is a given, and numerical integrators discretize its integral.
In ML, the setup is somewhat different, and the above conditions are typically ignored or assumed without verification.

\subsection{Numerical Integration of Ordinary Differential Equations}
\label{sec:overview-integrators}


In scientific computing, an \emph{initial value problem} (IVP) is an ODE problem of the form of Eq.~\eqref{eq:one}, where it is also specified that 
$x(t_0) = x_0$.
Solving an IVP involves computing the solution $x(t)$ using only knowledge of the function $\mathcal{R}(x,\theta)$, the specification of the parameters $\theta(t)$, and the initial condition $x(t_0) = x_0$.
To generalize our discussion (and simplify notation), we can discuss any time-dependent ODE by defining the function $f(x(t),t) = \mathcal{R}(x(t),\theta(t))$, where the form of $\mathcal{R}$ and the specific function $\theta(t)$ are assumed to be specified (to pose the problem fully).
Also, $x(t)$ can lie in any space, but (to simplify the discussion) we focus on the special case of a 1-tensor, i.e., $x(t)\in\mathbb{R}^n$.
If $f$ does not depend of $t$, other than the dependence on $x(t)$ (i.e., if $\theta(t)=\bar{\theta}$ is fixed in time), then Eq.~\eqref{eq:one} is called an \emph{autonomous} system; otherwise, it is a nonautonomous system. 
The existence and uniqueness of a solution $x(t)$ for a finite domain $t\in[0,T]$ can be proven under various assumptions, e.g., the Lipschitz-continuity of $f(x,t)$.
See~\cite{hirsch1974differential} for details. 
	
Numerical methods for solving IVPs involve constructing a sequence of successive estimates, $x_{0}, x_1,  \dots, x_{N_t}$, for $x(t)$, at successive time points $0=t_0< t_1 < \dots <t_{N_t}=T$.
(In numerical analysis, these points are called \emph{nodes} of the numerical integration scheme.) 
We  focus on uniformly spaced nodes, in which $t_k = k\Delta t$, where $\Delta t= T/N_t$ is the time step. 
(Using adaptive time steps, including those that have been used for ODE-nets~\cite{zhang2019anodev2, chen2018neural}, is of interest but is left for future work.)
The estimates $x_k \approx x(t_k)$ constructed through numerical integration schemes take advantage of the fact that any solution $x(t)$ must satisfy	%
\begin{equation}
	\label{eq:ode_int}
	x(t_{k+1}) = x(t_k) + \int_{t_k}^{t_{k+1}} f(x(t), t_k) \mathrm{d} t.
\end{equation}
	There are many numerical integration schemes, or numerical integrators, which are well understood, and there is a rich literature for how to balance computational complexity and approximation error~\cite{leveque2007finite}. 
	The (explicit) forward Euler method used in Eq.~\eqref{eqn:forward_euler_form} is the simplest approach, dating back to the work of Leonhard Euler in 1768.
	This method estimates the integral in Eq.~\eqref{eq:ode_int} using the left-sided rectangle rule
		\begin{equation}\label{eq:aghh}
		\int_{t_k}^{t_{k+1}} f(x(t), t_k) \mathrm{d} t \approx \Delta t \, f(x_k, t_k) .
		\end{equation}
	We obtain the familiar forward Euler update rule by plugging the left-sided rectangle rule into Eq.~\eqref{eq:ode_int},
	\begin{equation}\label{eq:FE}
		x_{k+1} = x_k + \Delta t \, f(x_k, t_k) .
	\end{equation}
	
	To evaluate the quality of such a numerical integration scheme, one can consider the global truncation error $E$ after predicting the solution at a final time $T=N_t\Delta t$ for a given time step $\Delta t$, defined as
	$E(\Delta t)=||x_{N_t}-x_{true}(N_t\Delta t)||$.
	That is, $E(\Delta t)$ is the error at the final time after performing all $N_t$ time steps, as opposed to the local trunctation error which only considers one step.
	The forward Euler update rule has first-order global truncation error, i.e., $E(\Delta t) = \mathcal{O}(\Delta {t})$. 
	Many more variations are possible. In general, one-step methods (which only use $x_k$) all take a loose form $x_{k+1}=x_k + \Delta t \,\, scheme(f,x_k,t_k\Delta t)$, where $scheme$ is some recurrence relation based on $f$. (We factor out the $\Delta t$ to highlight that there is always a leading-order dependence on $\Delta t$; there can be many more usages of $\Delta t$ inside of $scheme$.)
	One possible modification of single-stage forward Euler scheme leads to the explicit midpoint method that takes the~form 
	\begin{equation}\label{eq:midpoint}
		x_{k+1} = x_k + \Delta t \,\, f(x_k +  \frac{\Delta t}{2} f(x_k,t_k) , t_k + \frac{\Delta t}{2}  ).
	\end{equation}
	Unlike forward Euler, this scheme involves evaluating $f$ two times, which in turn leads to a more accurate IVP solver that has  second-order global error, i.e., $E(\Delta t) = \mathcal{O}(\Delta {t}^2)$. 
	Both the forward Euler method and the midpoint method are members of the family of Runge-Kutta methods, which were developed by Carl Runge and Wilhelm Kutta in the early 20th century. 
	Perhaps the most prominent member of this family, and the most widely used IVP solver, is the fourth-order Runge-Kutta (RK4) method, also called ``RK4-classic,'' which is an explicit method that has four intermediate \emph{stages}: 
	\begin{subequations}
		\begin{align}
		y_1 := & \,\,f(x_k, t_k) \\
		y_2 := & \,\,f\left(x_k+(\Delta t/2)y_1, t_k+\Delta t/2\right) \\
		y_3:= & \,\,f\left(x_k+(\Delta t/2) y_2, t_k+ \Delta t/2 \right) \\
		y_4:= & \,\,f\left( x_k + \Delta t y_3, t_k+\Delta t\right)  \\
		x_{k+1} = & \,\,x_k + \Delta t \left(y_1/{6} + y_2/{3}  + y_3/{3}  + y_4/{6}  \right).\label{eq:RK4}
		\end{align}
	\end{subequations}
	While RK4-classic involves evaluating $f$ four times sequentially (and so its computational cost is four times that of forward Euler), it's global error is fourth order, i.e.,	$E(\Delta t) = O(\Delta t^4)$, which decays as $\Delta t\to 0$ asymptotically \emph{much} faster than the $O(\Delta t)$ scaling of forward Euler. 
	In the following, we will consider RK4-classic, and we will also consider the variant of RK4 that uses the 3/8s integration rule, ``RK4-3/8,'' which also has fourth order error.
	
	Observe that the second terms on the right-hand side of Eq.~\eqref{eq:midpoint} and Eq.~\eqref{eq:RK4} are {\em also} scaled by $\Delta t$, just like forward Euler of Eq.~\eqref{eq:FE}.
	In fact, any explicit 1-step  integrator takes the form $x_{k+1} = x_k + \Delta t\,\, scheme(f,x_k,t_k,\Delta t)$ for some approximation scheme ($scheme$), with a leading dependence on $\Delta t$. Thus, in particular, \emph{all} of these explicit 1-step integrators are also similar to the syntactic algebraic form of a ResNet unit.

	\subsection{Learning Continuous Dynamical Systems with Discrete Models}
	\label{sec:overview-dynamical-setup}
	
	Here, we describe our results on learning a continuous dynamical system with discrete NN models.
	Importantly, rather than evaluating the quality of the learned model via training/test curves, which is typical in ML when one does not have access to the ``ground truth'' data, here we will evaluate the quality of the learned model---\emph{at the $\Delta t$ for which the model was trained, as well as for other values of $\Delta t$}---by examining both sample trajectories as well as the global error.
	
	\subsubsection{Basic Setup}
	
	To study the behavior of ODE-Nets in combination with different integrator schemes, we consider the prototypical nonlinear pendulum problem~\cite{hirsch1974differential}, where the angle $\rho(t)$ obeys the second-order~ODE
	\begin{equation}
	\label{eq:omegaddot}
	\ddot{\rho}(t) = 
	g \, \sin(\rho(t)),
	\end{equation}
	where $g=-9.81$ is a constant that represents the effect of gravity on the pendulum's mass, and $\rho=0$ at its lowest position.
	We set the pendulum's mass and radius to be equal to one.  
	We define $v(t) =\dot{\rho}(t)$ as the angular velocity, and we consider a stationary initial position with $v(0)=0$ and $\rho(0)=3\pi/4$.
	To construct an empirical training data set, we encode the state variables by $x(t)=[\rho(t),v(t)]$ at time steps $t_k = k \Delta t_{data}$ with a sampling rate $\Delta t_{data}$.
	For this particular nonlinear dynamical system, an analytical solution exists~\cite{belendez2007exact}, and this is used to provide the ``ground truth'' to evaluate different models (as opposed to using synthetic data produced by numerical solvers).
	Specifically, letting $K$ denote the complete elliptic integral of the first kind and $\mathrm{sn}$ denote the Jacobi elliptic function of the first kind, the solution of Eq.~\eqref{eq:omegaddot}
	is 
	\begin{equation}\label{eq:true}
	\rho(t)=2\arcsin\left( \sin({\rho}/{2}) ~\mathrm{sn}\Big(K\left(\sin^2({\rho_0}/{2})\right)-\omega_0t \, ; \, \sin^2({\rho}/{2})\Big) \right),
	\end{equation}
	where $\rho(0)=\rho_0$ is the initial condition and $\omega_0 = \sqrt{g}$ is the natural angular frequency.

	To learn this dynamical system from data, a common and widely applicable approach is to fit a model of the form
	\begin{equation}
	x_{k+1} = F(x_k,\Delta t; \hat{\theta}),
	\end{equation}
    where $F(\cdot)$ is a feed-forward model with trainable parameters.
	Here, we choose to construct feed forward models from a Neural ODE perspective. 
	At the core of these models is a neural network $G(x;\hat{\theta})$, which we desire to optimize to correspond to $\dot{x}$, the continuous-time dynamics.
	We construct the models' computational graphs by embedding $G$ into \emph{one step} of a fixed step-size integrator from the Runge Kutta family of numerical integrators,
	\begin{equation}
	 F(x_k) := \text{ODESolve}\left[G(x;\hat{\theta}),\, \, \Delta t, t_{start}=t_k,t_{end}=t_k+\Delta t] \right](x_k)  .
	\end{equation}	
	In particular, we use this general method to produce the graphs corresponding to steps of three different Runge-Kutta numerical integrator schemes: 
	\begin{itemize}
		\item 
		for the forward Euler method~\eqref{eq:FE}, the model is $F:= \mathtt{Euler}[G, x_k,\Delta t]= x_k+\Delta t G(x_k)$;
		
		\item
		for the explicit midpoint method~\eqref{eq:midpoint}, the model is $F:=\mathtt{Midpoint}[G,x_k,\Delta t]$; and
		
		\item
		for the fourth-order Runga Kutta (RK4-classic) method~\eqref{eq:RK4}, the model is $F:=\mathtt{RK4}[G,x_k,\Delta t]$.
	\end{itemize}
	For all three cases, $G: \mathbb{R}^2 \rightarrow \mathbb{R}^2 $ is chosen to be a simple two-layer NN, 
	with $D=50$ hidden units
	\begin{equation}
	G(x; \hat{\theta}) =  \hat{A} \, \tanh\left( \hat{W} x + \hat{b}\right),
	\end{equation}
	where $\hat{\theta}=\{\hat{A},\hat{W},\hat{b}\}$ are the learnable parameters of the model.
	We denote these models as ODE-Net(Euler), ODE-Net(Midpoint), and ODE-Net(RK4),
	and we use torchdiffeq~\cite{chen2018neural} to implement them. 
	These models are essentially one-step feed-forward models, whose entire computational graphs are the units illustrated in Fig.~\ref{fig:resnet}, with $G$ in place of $\mathcal{R}$.
	Note also that the ODE-Net(Euler) model, i.e., $F:= x_k+\Delta t G(x_k)$, resembles the discrete ResNet unit in Eq.~\eqref{eqn:forward_euler_form}. This form also resembles a general class of temporal difference models which predate the interpretations of ODE-Nets or residual networks~\cite{sutton1988learning,jordan1990learning}.
		
Given this setup, we use the Adam optimization algorithm~\cite{kingma2014adam} to minimize the loss of the prediction error of one step in~time,
\begin{equation}
\min_\theta \frac{1}{N_t} \sum_{k=0}^{N_{data}} \left\| F(x_k,\Delta t; \hat{\theta} ) -  x_{k+1} \right\|_2^2,
\end{equation}
using graph-based back-propagation to compute the gradients.

\subsubsection{Model Quality via Sample Trajectories for Different Time Steps}

	\begin{figure}[t] 
		\centering
		\begin{subfigure}[t]{0.9\textwidth}
			\centering
			\begin{overpic}[width=1\textwidth]{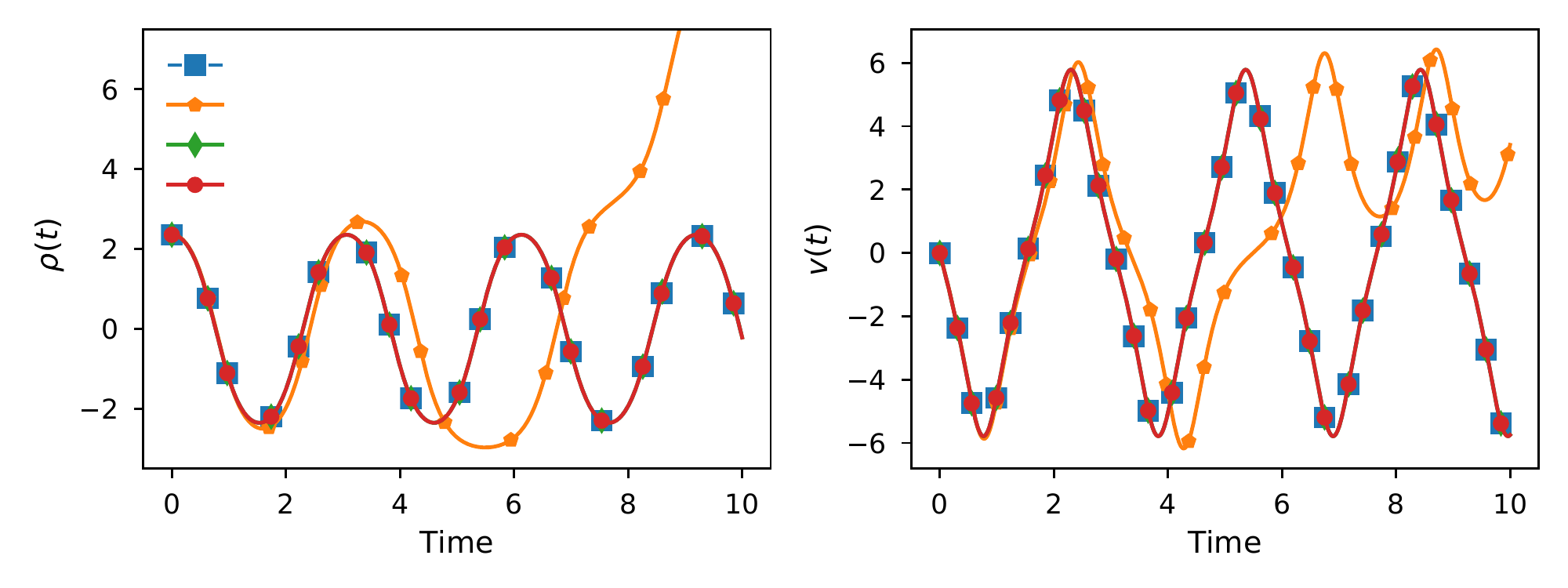}
			\put(15.5,32.5){\footnotesize Ground Truth}				
			\put(15.5,30.0){\footnotesize Numerical Euler}
			\put(15.5,27.5){\footnotesize ODE-Net(Euler)}
			\put(15.5,25){\footnotesize ODE-Net(RK4)}			
			\end{overpic}\vspace{-0.3cm}		
			\caption{$\Delta t=\Delta t_{data}$}
		\end{subfigure}\vspace{+0.2cm}
		
		\begin{subfigure}[t]{0.85\textwidth}
			\centering
			\begin{overpic}[width=1\textwidth]{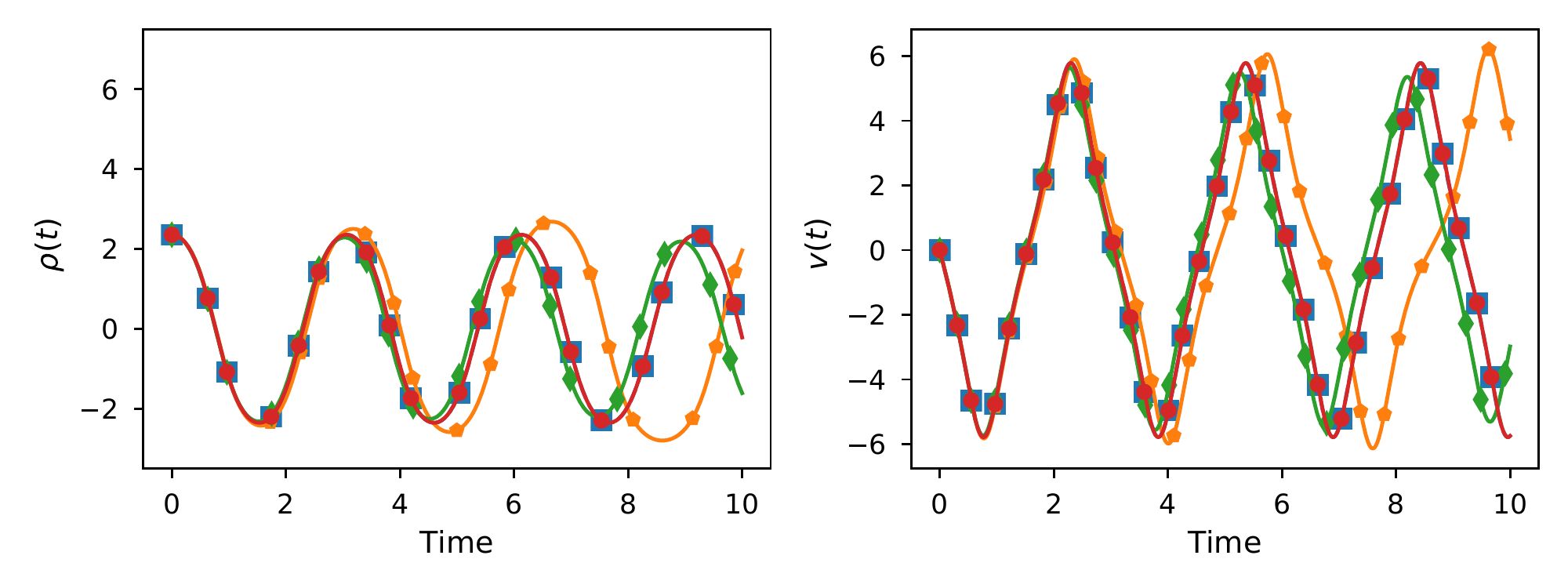}		
			\end{overpic}\vspace{-0.3cm}			
			\caption{
				$\Delta t=0.5\Delta t_{data}$
			}
		\end{subfigure}	
		
		\caption{In (a), we show trajectories for models at the dataset's sampling rate; and in (b), we show trajectories generated by models that attempt to predict the time series at twice the sampling rate. 
		(Some of the models have global error $\approx 10^{-4}$, meaning in particular that the curves of the ODE-Net(Euler) and ODE-Net(RK4) both lay on top of the ground truth curve in (a).) 
		A model that approximates a continuous differential equation will become more accurate when the time step $\Delta t$ is decreased.
		However, ODE-Net(Euler) becomes less accurate and diverges significantly, indicating that it has overfit the sampled data and the discrete time step. 
		(Note that the plot markers are not drawn on every time step; the data is sampled at $\Delta t=0.01$ on top and $\Delta t=0.005$ on bottom.)
		}
		\label{fig:pendulum_trajectories}
	\end{figure}

We first study model quality by examining  sample trajectories.
In Fig.~\ref{fig:pendulum_trajectories}, we show  trajectories, i.e., $\rho(t)$ and $v(t)=\dot{\rho}(t)$, that are learned by the ODE-Net(Euler) and ODE-Net(RK4) models. 
(We do not show trajectories for ODE-Net(Midpoint), because they are indistinguishable from  ODE-Net(RK4) in these plots.)
We compare them to  the  ``ground truth'' trajectory, given by Eq.~\eqref{eq:true}.
We also compare them to the approximate solution for Eq.~\eqref{eq:omegaddot} that is obtained by applying the forward Euler numerical integrator to Eq.~\eqref{eq:FE} in the traditional way, i.e., we  iterate $x_{k+1} = x_k + \Delta t [\dot{\rho}(t_k),\dot{v}(t_k)]$.
This approach does not have access the the training data, and we denote it by Numerical~Euler.

In the top row of Fig.~\ref{fig:pendulum_trajectories}, we show trajectories with $\Delta t=\Delta t_{data}$; that is, when the trajectories are constructed using a time step that is identical to the data's sampling rate. 
As expected, we observe that the discrete Numerical Euler is an inaccurate and unstable representation for this continuous differential equation.
Recall that the forward Euler scheme is an explicit method, i.e., the state variables at a later time are directly calculated from the prior state variables.
In contrast, for this $\Delta t=\Delta t_{data}$, both ODE-Nets are accurately fit to the ground truth trajectory.

In the bottom row of Fig.~\ref{fig:pendulum_trajectories}, we depict trajectories when the time step is decreased to $\Delta t=0.5\Delta t_{data}$. That is, the models were trained at $\Delta t_{data}$, but we are plugging in a new (smaller) $\Delta t$ to perform inference at a higher sampling rate (i.e., smaller time step) to effectively interpolate the time series.
As expected, the trajectory for 
Numerical Euler improves in accuracy when the time step is decreased (but it still eventually diverges, not shown).
In contrast, observe that ODE-Net(Euler) is \emph{less} accurate when the time step is decreased. 
That is, this discrete NN model is ``overfit'' to the training time series data, for the particular time step $\Delta t=\Delta t_{data}$, and therefore it does \emph{not} learn a continuous system.
On the other hand, ODE-Net(RK4) remains accurate even after $\Delta t$ is changed (as does ODE-Net(Midpoint), which is not shown here). In turn, this suggests that these models are better able to learn the continuous system.

\subsubsection{Model Quality via Global Error for Different Time Steps}

We next examine model quality by examining the global error.
A continuous-in-time model (or, more precisely, a parameterized sequence of discrete models, such as a numerical integrator, that approximates a continuous model) should become more accurate as the time step $\Delta t$ decreases.
This can be quantified by a decrease in the global error, when the fully calculated trajectory is compared to the true solution at final time $T=N_t\Delta t$ for a given time step $\Delta t$,
\begin{equation}
	    E(\Delta t) = \| x_{true}(T) - F(F(\dots F(x_0;\hat{\theta},\Delta t)\dots;\hat{\theta},\Delta t);\hat{\theta},\Delta t)\|_2^2  ,
\end{equation}
where $F(\dots F(x;\hat{\theta},\Delta t);\dots \hat{\theta},\Delta t)$ refers to a chain of feed-forward recurrence, i.e., applying the model to its outputs to evaluate $N_t$ steps with the pre-trained parameters $\hat{\theta}$.
It is well-known that $E(\Delta t) =\mathcal{O}(\Delta t)$ for forward Euler, 	$ E(\Delta t) = \mathcal{O}(\Delta t^2)$ for midpoint, and $ E(\Delta t) = \mathcal{O}(\Delta t^4)$ for RK4. 
If one plots $E(\Delta t)$ versus $\Delta t$ in logarithmic axes, then 
\begin{equation}
	\log(E(\Delta t)) \propto  r \log(\Delta t),
\end{equation}
i.e., an integrator with order-$r$ error corresponds to a line with slope $r$.

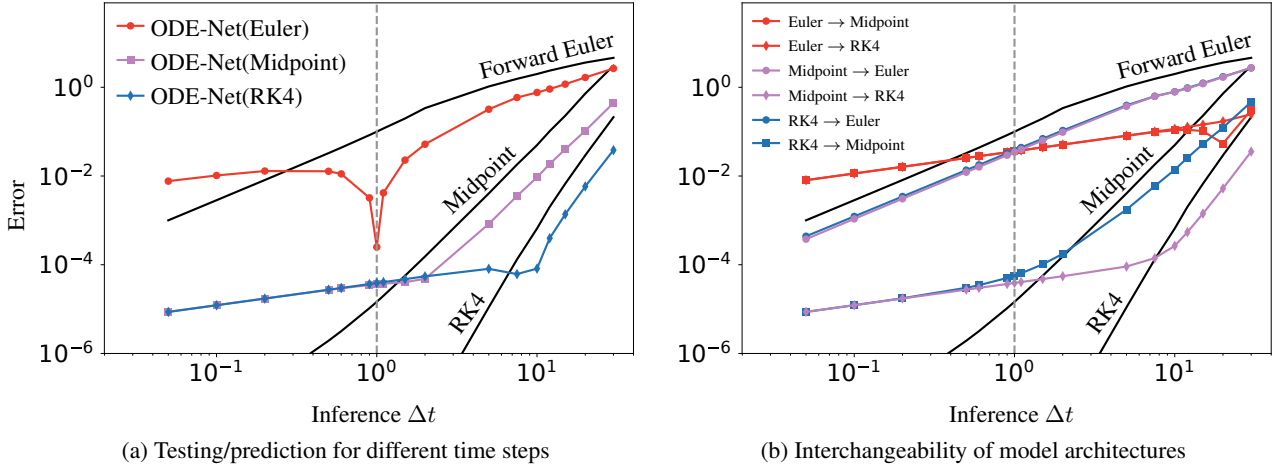
\begin{figure}[t] 
		\centering
		
		\begin{subfigure}[t]{0.49\textwidth}
			\centering
			\begin{overpic}[width=1\textwidth]{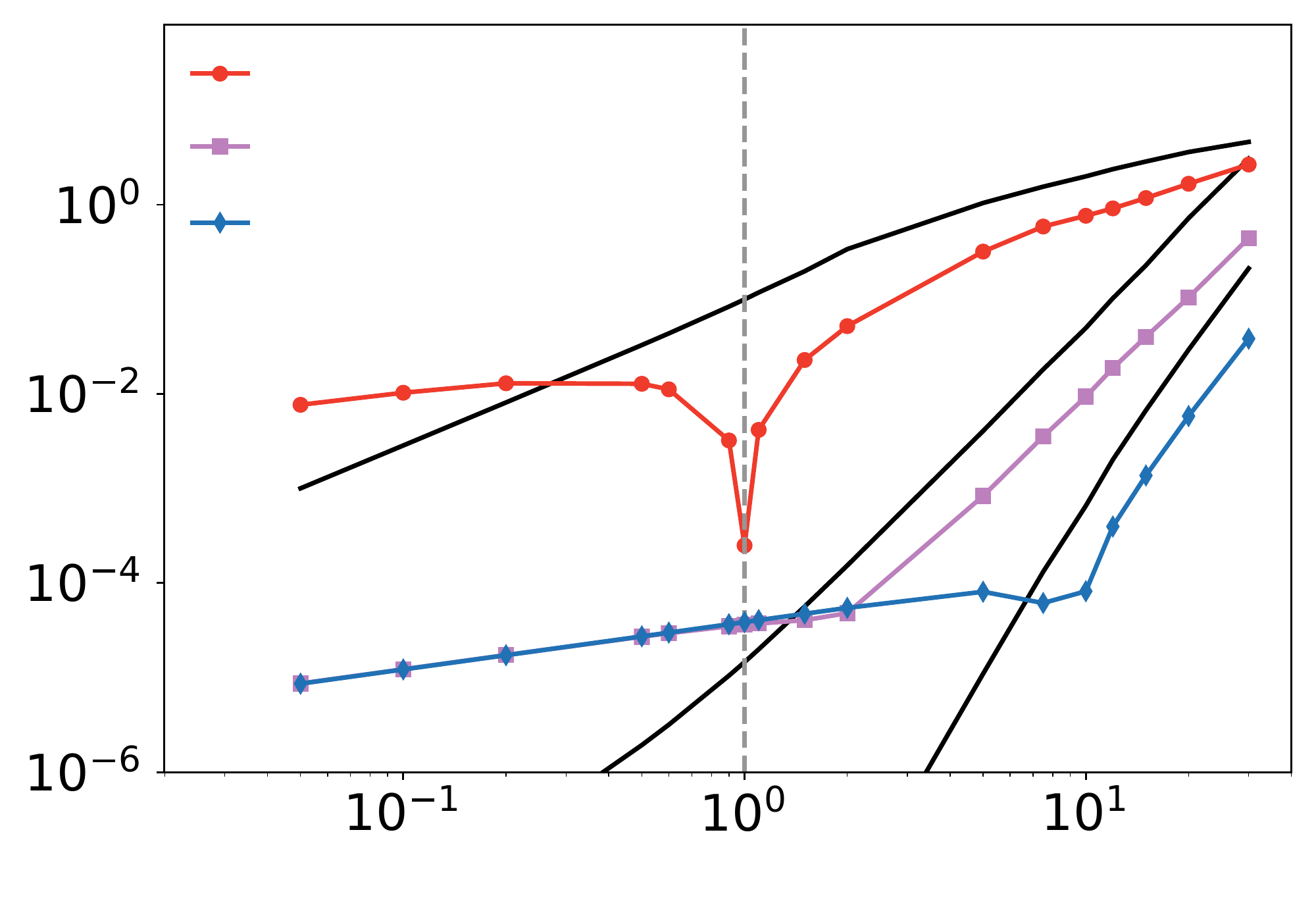}
				\put(-3,30){\footnotesize \rotatebox{90}{Error}}
				\put(46,-1.0){\footnotesize Inference $\Delta t$}  	
				
				\put(20,61.5){\footnotesize ODE-Net(Euler)}
				\put(20,56.0){\footnotesize ODE-Net(Midpoint)}
				\put(20,50.5){\footnotesize ODE-Net(RK4)}		
				
				\put(73,54.5){\rotatebox{14}{\footnotesize Forward Euler}}			
				\put(67,32.5){\rotatebox{43}{\footnotesize Midpoint}}	
				\put(68,13){\rotatebox{54}{\footnotesize RK4}} 
				
			\end{overpic}\vspace{+0.1cm}	
			\caption{Testing/prediction for different time steps}
			\label{fig:pendulum_theta_convergence_a}
			
		\end{subfigure}
		~
		\begin{subfigure}[t]{0.49\textwidth}
			\centering
			\begin{overpic}[width=1\textwidth]{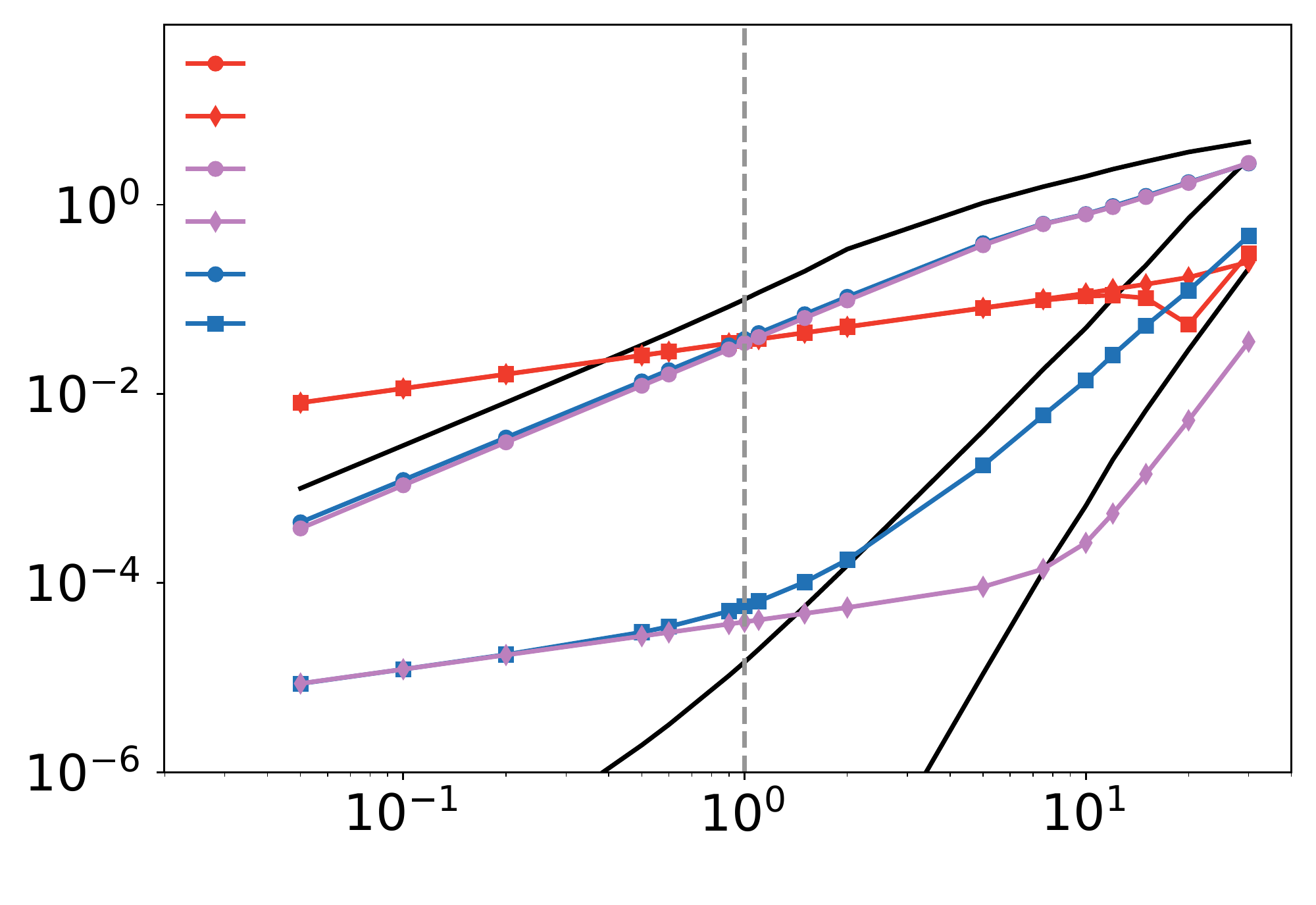} 
				\put(46,-1.0){\footnotesize Inference $\Delta t$}  	
				
				\put(20,63){\tiny Euler $\rightarrow$ Midpoint}
				\put(20,59){\tiny Euler $\rightarrow$ RK4}

				\put(20,55){\tiny Midpoint  $\rightarrow$  Euler}		
				\put(20,51){\tiny Midpoint  $\rightarrow$  RK4}		
				
				\put(20,47){\tiny RK4  $\rightarrow$  Euler}		
				\put(20,43){\tiny RK4  $\rightarrow$  Midpoint}	
				
				\put(73,54.5){\rotatebox{14}{\footnotesize Forward Euler}}			
				\put(67,32.5){\rotatebox{43}{\footnotesize Midpoint}}	
				\put(68,13){\rotatebox{54}{\footnotesize RK4}} 	
				
			\end{overpic}\vspace{+0.1cm}			
			\caption{Interchangeability of model architectures}
			\label{fig:pendulum_theta_convergence_b}
			
		\end{subfigure}   
		
		\caption{
			Convergence results for ODE-Nets trained on data sampled at rate $\Delta t = 1$; and (black curves)  the numerical integrators. (a) The ODE-Net based on forward Euler has a large ``dip'' at $\Delta t=1$, revealing it is overfit to the training data; but models based on higher-order integrators are more accurate and robust.
			(b) The midpoint and RK4 models can be robustly substituted into  other integrators, whereas the forward Euler model cannot be substituted.
		}
		\label{fig:pendulum_theta_convergence}
\end{figure}

In Fig.~\ref{fig:pendulum_theta_convergence}, we study the error $E(\Delta t)$ for the trajectories of ODE-Nets as we vary $\Delta t$, and we compare the results to the convergence behavior of numerical integrators.
The solid black lines denote $E(\Delta t)$ for the numerical integrators, and their slopes are approximately $r=1$, $2$ and $4$. These lines provide baselines that we use to compare to the errors of the ODE-Nets.
The curves show that the errors $E(\Delta t)$ of the ODE-Nets behave very differently from those of the numerical~integrators.

More concretely, in Fig.~\ref{fig:pendulum_theta_convergence_a}, we see that the error for ODE-Net(Euler) has a large ``dip'' at $\Delta t = \Delta t_{data}$. 
This shows two related things: first, at this precise time step,  ODE-Net(Euler) is \emph{significantly} more accurate than would be suggested by a forward Euler interpretation; and second,  ODE-Net(Euler) is strongly overfit to the data that is sampled at rate $\Delta t_{data}$. 
That is, ODE-Net(Euler) is a good \emph{discrete} model, without changing $\Delta t$ or the computational graph, but it is \emph{not} meaningfully a continuous model or a discrete approximation to a continuous model, since it is \emph{extremely} sensitive to perturbations to $\Delta t$.
In contrast, the error curves in Fig.~\ref{fig:pendulum_theta_convergence_a} for ODE-Nets(Midpoint) and ODE-Nets(RK4) do not have a dip; they are robust to changes in $\Delta t$. 
That is, they are good discrete models (since they have small errors), and they are also good (discrete approximations to) continuous models.
This is consistent with our results in Fig.~\ref{fig:pendulum_trajectories}.
Observe also that, for large $\Delta t\gtrsim 10$, the error of the ODE-Nets behaves similar to their respective integrators, i.e., their slopes match the baseline slopes. 
However, the optimization process ``leaves behind'' some error $e_{opt}=\|x_{t+1}-F(x_t;\hat{\theta},\Delta t_{data})\|$, and this affects how well $G$ itself represents the dynamics. 
The error of the learned models is composed both of the truncation error and the optimization error.
There is then a threshold at $E(\Delta t)=e_{opt}$ where in the $\Delta t\rightarrow 0$ limit the cannot continue to approach zero at the same rate. 
However, within the bounds of the training error, the learned $G$ behaves similarly as the ODE.

In Fig.~\ref{fig:pendulum_theta_convergence_b}, we illustrate another important aspect of learning an ODE.
Given that $G$ accurately approximates an ODE, it should be possible to interchange its computational graph with another one that takes the form of one for another numerical integrator scheme.
That is because different discrete numerical integrators applied to the same ODE will all approximately  solve that same  ODE. 
As we can see, when the $G$ obtained by a computational graph that represents the RK4 method is inserted into a graph that represents the Midpoint method, the results are consistent with the expectations for the baseline solution obtained by the numerical Midpoint scheme.
The same holds vice versa for Midpoint, where the new model is consistent with the baseline RK4 numerical method.
In this case, we show that it is actually possible to improve the inference accuracy by plugging into a higher order integrator than was used for training.
When looking at the case of plugging in the $G$ learned using Midpoint into an RK4 integrator, the error is improved by orders of magnitude for $\Delta t>2 \Delta t_{data}$.
However, the rate of convergence is limited when the truncation error is below the optimization error $e_{opt}$, similar to the unmodified graphs on the right panel.
In contrast, performing this substitution using $G$ learned with forward Euler into either of the higher order integrators results in orders of magnitude higher-order error, i.e., much \emph{worse} performance, than the original model for any time step.
Even without changing the time step away from $\Delta t_{data}$, neither of the two transformed models is accurate.

\subsection{Implications for Continuous-time Machine Learning Models}

When a numerical integrator is used in scientific computing, we start with a differential equation, and we formulate a discrete model which can produce discrete data. 
Learning a model using a data-driven approach, however, works in the opposite direction, i.e., we fit a discrete model to a collection of discrete data points.
Even if the models are algebraically the same, analogous components of a learned model do not necessarily correspond to the continuous kernel of a numerical integrator.
Our results in Fig.~\ref{fig:pendulum_trajectories} and Fig.~\ref{fig:pendulum_theta_convergence} on learning a model for data generated from the non-linear pendulum problem can be summarized as follows:
\begin{itemize}
	\item  
	A ResNet model, i.e., an ODE-Net that uses the forward Euler integration scheme, is much more accurate than we would expect if it were truly a forward Euler discretization of a continuous ODE.
	\item 
	The higher-than-expected accuracy is not robust to changes to $\Delta t$.
	The discrete residual model has discretization error comparable to the noise in the data, and it overfits to the sample rate of the time series data. 
	\item
	ODE-Nets that use higher-order architectures corresponding to more sophisticated numerical integration schemes, such as the Midpoint method or the RK4 method, have much lower numerical discretization error, and they do exhibit the properties we would expect of a discrete approximation to a continuous ODE.
\end{itemize}
These results demonstrate that the ODE-Net(Euler) model cannot be interpreted as a forward Euler discretization of the differential equation, despite having a similar algebraic form.
That is, the analogy of a ResNet to forward Euler is superficial and at best incomplete. 
These results also demonstrate that ODE-Net(Midpoint) and ODE-Net(RK4) can meaningfully be interpreted as a discrete approximation to a continuous dynamical system.
We next use these insights to develop continuous-in-depth NN models more generally.

\section{Continuous-in-Depth Neural Networks}
\label{sxn:continuous-in-depth}

In this section, we introduce \CoolNet{}, a continuous-in-depth generalization of deep ResNets.
In \CoolNet{}, a discrete integer-valued depth indexing into a particular residual unit is replaced by continuous real-valued time evaluating a differential amount of computation.
Using our insights from Sec.~\ref{sec:overview} on learning (or, depending on the specific model, not learning) a continuous dynamical system with a discrete numerical integrator, we show that \CoolNet{} models can be trained to meaningfully correspond a continuous dynamical system.

To express ``deep'' computations in a tractable way, the weights for each differential step of the residual module are computed from a basis function representation in time (depth).
Steps of the corresponding computational graph correspond to stages of a numerical integrator which are assigned weights through the basis functions.
Importantly, this approach enables us to decouple the computational graph from the model parameters, a property which does \emph{not} hold for traditional discrete NNs but which is crucial for being a numerical integrator in a meaningful sense.
In addition to enabling one to learn a continuous dynamical system, this decoupling also has many benefits for more traditional ML applications.
In particular, similar to the high-order integrators in Sec.~\ref{sec:overview}, \CoolNet{} architectures exhibit manifestation invariance: they are able to freely manifest many different computational graphs, with different depths or integrator structures, without the need to alter the learned weights.

\subsection{Design Principles for \CoolNets{}}
\label{sxn:design_principles}

\begin{figure}[!t]
	\centering
	\includegraphics[width=0.95\linewidth]{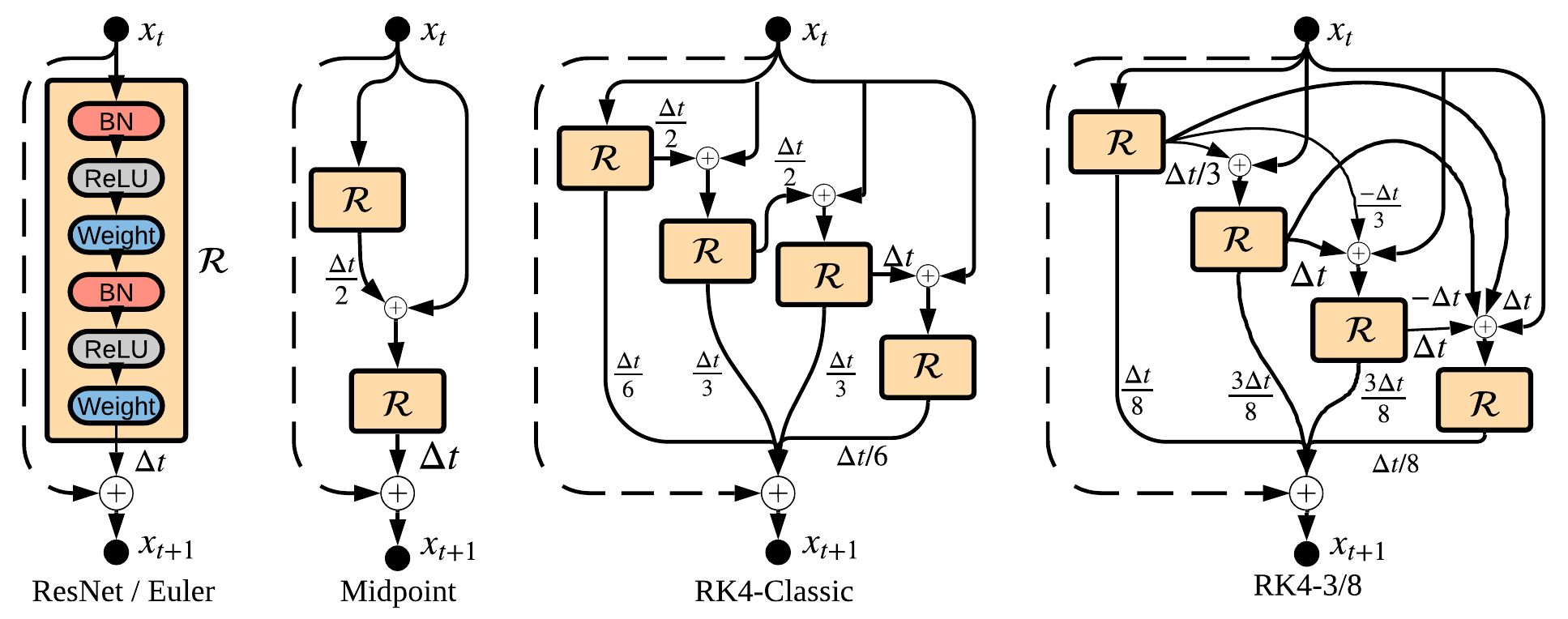}
	\caption{
		We build interchangeable \CoolNet{} units  by repeating a ResNet unit $\mathcal{R}$ inside of  the architecture of higher-order numerical integrators: (left) forward Euler, given by Eq.~\eqref{eqn:forward_euler_form}, which recovers the ResNet unit; (next) midpoint;
		(next) RK4-classic, given by Eq.~\eqref{eq:RK4}; and (right) RK4-3/8.
		The labels on the edges (e.g., $\Delta t/2$), indicate the constants that are needed by the different integrators.}
	\label{fig:resnet}
\end{figure}

In traditional discrete ResNets, the residual module $\mathcal{R}(x_k,\theta_k)$ is a shallow convolutional network unit involving a sequence of batch normalization, activation function, and convolution. 
Various configurations are possible, e.g., the configuration illustrated in the left panel of Fig.~\ref{fig:resnet}. Each residual unit $k$ has its own parameters $\theta_k$, and so the number of parameters grows with the network depth. 

We propose a generalization that is based on an ODE:
\begin{equation}
\label{eq:refinenet_split_ode}
\dot{x}(t) = \epsilon \, \mathcal{R}\left(x(t),\theta(t;\hat{\theta})\right)  ,
\end{equation}
where the layers' weights are replaced by a weight function $\theta(t)$ that is evaluated at different points in time $t \in[0,T]$, and where  $\epsilon$ is a tuning hyperparameter that adjusts the time scaling. The update map $\mathcal{R}$ can be any residual module, such as any which have previously studied in the literature, including, e.g., the original skinny modules with or without bottlenecks \cite{he2016deep}, the wide variants \cite{zagoruyko2016wide}, or aggregated modules \cite{xie2017aggregated}.
By itself, $\mathcal{R}$ is a pure function of $x$ and $\theta$, where $x$ is one instance of the intermediate state, and $\theta$ is one instance of the parameters needed by the sub-components of $\mathcal{R}$. 
That is, in our formulation, $\theta$ is an input to $\mathcal{R}$, instead of a tensor ``owned'' by a given unit. Multiple units (i.e., instances of $\mathcal{R}$ in the graph) can be passed in the same value of $\theta$; and, with higher order graphs, multiple values of $\theta$ can be used inside of one unit with multiple instances of $\mathcal{R}$.

Integration of Eq.~\eqref{eq:refinenet_split_ode} is the mechanism for feed-forward computation of a continuous-in-depth network. 
\CoolNets{} provide building blocks that can be incorporated into deeper architectures, wherein each OdeBlock module learns its own continuous function. 
The input-output relation of the network element is:
\begin{align}
\label{eq:odeblock_inout}
x_{out} &=  x_{in}+\int_0^T \mathcal{R}\left(x(t),\theta(t)\right) \,\mathrm{d}t \\
&= \mathtt{OdeBlock}\left[\mathcal{R},\theta,scheme,\Delta t , t\in[0,T]\right](x_{in}).
\label{eq:odeblock_inout_2}
\end{align}
These can be combined into very deep architectures in the same way as ResNet blocks. 
The total ``time'' $T$ is a fixed hyperparameter for a given network (corresponding to continuous depth), and varying it yields the inverse effect as varying $\epsilon$. 
Without loss of generality, we choose to fix $T=1$, and thus we only treat $\epsilon$ (corresponding to inverse depth) as a hyperparameter when implementing \CoolNet{}.%
\footnote{Some existing ResNets also have an $\epsilon$ tuning parameter that can be used to control the magnitude of the update map. This $\epsilon_{\text{resnet}}$ is also part of the gauge formed with $T$ and $\epsilon_{ode}$.  Thus, for \CoolNets{}, we only have one $\epsilon$ parameter.}

To specify a \CoolNet{} model, as given in Eq.~\eqref{eq:odeblock_inout_2}, we need two things: a specification of the layers in $\mathcal{R}$, such as $\{BN,ReLU,Conv2d,BN,ReLU,Conv2d\}$; and a specification of $\theta(t;\hat{\theta})$, a function continuous in $t$ with trainable parameters $\hat{\theta}$.
We describe the particulars of how $\theta(t)$ is represented by discrete parameters $\hat{\theta}$ below in Sec.~\ref{sxn:construction_of_basis_functions}. 
Given an instance of parameters $\hat{\theta}$, there are many ways to evaluate Eq.~\eqref{eq:odeblock_inout_2}. 
Before the \CoolNet{} prediction is to be evaluated for a given set of inputs, a choice must be made for the numerical integration scheme, $scheme$, and for the time step (depth) $\Delta t=1/N_t$.
From the discrete deep NN perspective, these choices effect a computational graph manifestation.
The choice of $scheme$ dictates which of the units in Fig.~\ref{fig:resnet} are chosen, and each of these prescribes a different way to nest the residual $\mathcal{R}$. 
Because every numerical integrator provably solves the same problem, the weights shown in Fig.~\ref{fig:resnet} provide consistent ways of rearranging $\mathcal{R}$ around the skip connection (but, of course, with different ramifications for numerical accuracy and numerical stability).
The choice of $N_t$ dictates how many times the unit is repeated to make the deep NN. 
Each invocation of $\mathcal{R}$ is assigned a different time $t\in[0,T]$ according to the numerical integration procedure. 
The basis functions within $\theta(t)$, described in Sec.~\ref{sxn:construction_of_basis_functions}, dictate how each invocation of $\mathcal{R}$ inside of this graph maps to particular entries in $\hat{\theta}$. 

To summarize, given the continuous-in-depth model in Eq.~\eqref{eq:odeblock_inout_2}, the ($scheme$,$N_t$) pair provides instructions for manifesting an entire family of equivalent discrete deep NN graphs, as illustrated in Fig.~\ref{fig:\CoolNet_graph}, all of which evaluate the same \CoolNet{} model, prescribed by only $\mathcal{R}$ and $\theta(t)$.

\subsection{Construction of Basis Functions}
\label{sxn:construction_of_basis_functions}

In a traditional discrete ResNet, with layers $k=0,1,\dots,N_t$, each step evaluates the same function $\mathcal{R}(x_k,\hat{\theta}_k)$, with a series of different parameters $\hat{\theta}_1, \hat{\theta}_2,\dots \hat{\theta}_{N_t}$.
These are indexed by $k$, and these are different for each discrete layer.%
\footnote{The notation $\hat{\theta}_k$ refers to all of the weight tensors used by each of the network elements of the $k$th residual unit, e.g., $\hat{\theta}_k:=\left\{\mathtt{conv2d:}w_{ijkl},\,\mathtt{conv2d:}b_{ij},\,\mathtt{bn1:}\mu\,\dots \right\}$. Similarly, the notation $\theta(t)$ refers to all of the individual weight functions used to construct each tensor entry of the network elements, i.e., $\theta(t):=\left\{\mathtt{conv1:}w_{ijkl}(t),\,\mathtt{conv1:}b_{ij}(t),\,\mathtt{bn1:}\mu(t),\dots \right\}$.} 
To construct our continuous-in-depth generalization of ResNet, we want a parameter function, $\theta(t)$, which varies continuously (in some sense) in time, i.e., continuously in depth.
Clearly, for any choice $N_t$ of discrete depth, this will yield a different mapping of parameters to each layer, but our construction will ensure that any mapping results in a valid approximation to the continuous function $\theta(t)$.

To accomplish this, we present an elementary function construction that introduces the concept of time (noting, of course, that more advanced constructions are possible).
First, we associate the parameters at each step $k$ to point in time $t_k\rightarrow \theta_k$, where the time points are an equally spaced grid $t_k=k\Delta t$, where $\Delta t = T/N_t$. 
Then, since these parameters have been defined only at specific discrete points at time, to ``fill in the gaps'' and obtain a continuous function that is defined on the continuous time domain $t\in [0,T]$, we say that each point on this grid represents one interval in time with a constant value of $\theta_k$.
Thus, the ResNet weights can be represented by the following function $\theta(t)$,
\begin{equation}
\label{eq:piecewisetheta}
\theta(t) =\begin{cases}
\hat{\theta}_1 &, \,\, t\in [0, \Delta t) \\
\hat{\theta}_2 &, \,\, t\in [\Delta t, 2 \Delta t) \\
... \\
\hat{\theta}_{N_t} &, \,\, t\in [T-\Delta t, T]  .
\end{cases}
\end{equation}
When applying the forward Euler integration scheme, this function picks out $\theta_k$ for step $k$ starting at $t_k$, but it can also be evaluated with different integrators and with smaller or larger values of $\Delta t$.

Importantly, the functional representation chosen for $\theta(t)$ can be generalized further from Eq.~\eqref{eq:piecewisetheta} by introducing the notion of continuous-in-time basis functions.
In \CoolNet{}, all weights within an OdeBlock are be obtained by evaluating a time-dependent tensor $\theta(t)$ at some time $t$.  
The entries of $\theta(t)$ can be represented as a linear combination of $M$ basis functions $\phi(t)$,
\begin{equation}\label{eq:weight}
\theta(t;\hat{\theta}) = \sum_{\beta=1}^{M} \phi^\beta(t) \hat{\theta}^\beta.
\end{equation}
Given this basis function representation of $\theta(t;\hat{\theta})$, the components $\{\hat{\theta}^\beta\}_{\beta=1}^{M}$ of this functional basis are the weights that are learned during the training process.
There are $M$ coefficients for each tensor entry in $\theta(t)$ that is required by the specification of $\mathcal{R}$.
For example, the weight function $w_{ijkl}(t)\in \theta(t)$ for one of the $16\times16\times3\times3$ $\mathtt{conv2d}$ blocks has $M$ coefficients, so its parameters are a $M\times16\times16\times3\times3$ tensor, $w_{ijkl}(t)=\sum \phi^\beta(t)\,w_{ijkl}^\beta$.
In principle, one can choose any set of basis functions, and their tradeoffs should be explored in future work.
If one wants to recover the function implied by Eq.~\eqref{eq:piecewisetheta}, then one can choose an orthogonal basis of piecewise-constant ``indicator'' functions: 
\begin{equation}\label{eq:piece_const}
    \phi^\beta(t) = 
    \begin{cases}
    1, & t \in [(\beta-1) \Delta t,\, \beta\Delta t]\\
    0, & \mathrm{otherwise} .
    \end{cases}
\end{equation}
Here, each function $\phi^\beta(t)$ covers a nonoverlapping interval of the time domain $[0,T]$ and $\Delta t = T/M$.
While $M$ does not need to equal $N_t$, if one chooses $M=N_t$, then the coefficients $\hat{\theta}^\beta$ correspond to the original ResNet parameters $\hat{\theta}_k$.

\subsection{Incorporation of OdeBlocks into Problem-Specific Architectures}

OdeBlocks are a generalization of ResNet blocks, and they can provide a 1-to-1 replacement in existing architectures designed for any task.
The set of possible discrete computational graphs that can be generated by OdeBlocks (using the principles outlined in Sec.~\ref{sxn:design_principles} and Sec.~\ref{sxn:construction_of_basis_functions}) includes ResNet blocks. Thus, \CoolNets{} can generate exactly the same architecture for any problem as ResNet models.

\begin{figure}[!t]
\centering
\includegraphics[width=0.9\textwidth]{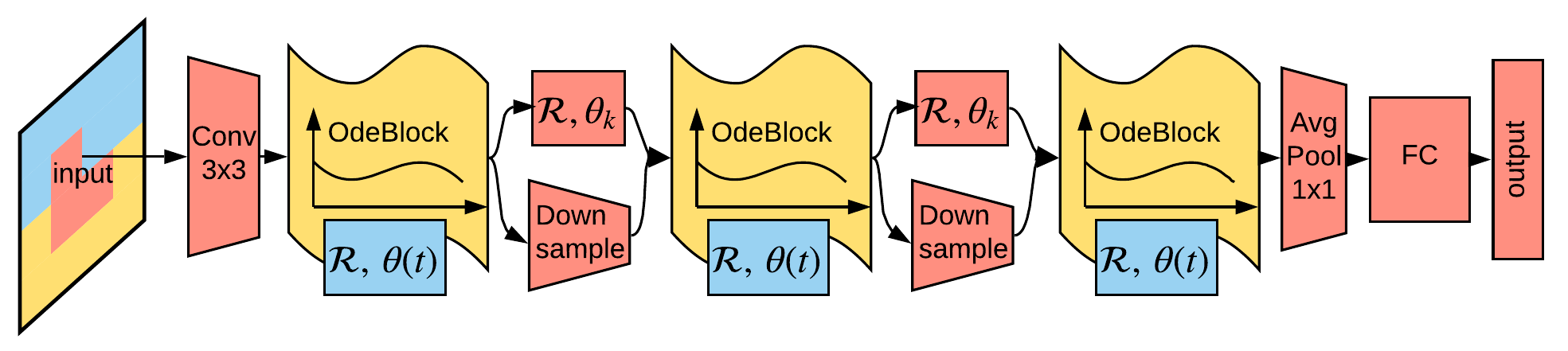}
\caption{%
\CoolNet{}  architecture, in which OdeBlocks provide a 1-to-1 replacement for ResNet blocks. 
OdeBlocks have computational graphs  that are based on numerical integrators. 
Like ResNet blocks, OdeBlocks can be assembled into a complex architecture alongside standard NN components (see the red components). 
Choosing the simplest integrator, forward Euler, exactly recovers a deep ResNet architecture.
Choosing higher-order integrators, e.g., RK4, allows OdeBlocks to learn differentiable dynamical systems and to construct continuous-in-depth NNs.
}
\label{fig:odenet}
\end{figure}

For instance, Fig.~\ref{fig:odenet} depicts a \CoolNet{} architecture that generalizes a standard ResNet for image recognition.
The architecture involves a sequence of three OdeBlocks in place of the residual-blocks of \cite{veit2016residual}, and each encodes a different continuous-in-depth transformation.
In this design, each OdeBlock has a different shape to reflect different image sizes and different numbers of filters after down-sampling.
The continuous model can be trained with different numerical integrators and different time steps, yielding a different discrete models in each case. 
For the following configurations, an OdeBlock yields the same computational graph as a ResNet block: forward Euler integration with $N_t$ steps; the same number $M=N_t$ of piecewise basis functions for $\theta$; and use the time-scaling $\epsilon=N_t$ with $\Delta t = T/N_t$.
For other configurations, an OdeBlock yields a different computational graph.

In \CoolNet{}, the OdeBlock components are assembled alongside the usual discrete NN components to form a complete network. 
For image recognition, the additional components are a linear convolutional layer at the beginning, the average-pool and fully-connected classifier at the end, and two ``stitching'' down-sampling residual units between the OdeBlocks. 
(For other tasks, different components could be used.)
The assembly is similar to how the residual network blocks are connected together using network segments that are not strictly residual units.
Continuity is broken when the signal is reshaped and pooled, and a ResNet downsample-and-skip unit is used to stitch together between different continuous domains defined by OdeBlocks.
Each of the three OdeBlocks have their own  $\mathcal{R}$ and $\theta(t)$; and each is connected as discrete network elements within the complete architecture via their inputs, $x_{in}=x(0)$, and outputs, $x_{out}=x(T)$.
The specifics of this architecture, such as the number of channels, can be chosen the same way as for discrete ResNet configurations. 
Specific values for our empirical studies are discussed in Sec.~\ref{sxn:refinenet-empirical}.
In this paper, the three OdeBlocks are always assigned the same $N_t$, $M$, and scheme to create uniform time steps and time partitions, but other nonuniform choices can be considered in future work.
See Appendix~\ref{sec:app-additional-details} for additional details.

\subsection{\CoolNets{} can be Seamlessly and Adaptively Refined}

Continuous-in-depth NNs are independent of any particular choice of discrete computational graph or parameter representation, yielding the property of manifestation invariance. 
This property can be exploited to create new strategies to systematically alter the discrete NN graph and weights during training, testing, or deployment of a \CoolNet{}.
Because \CoolNet{} is constructed to be a dynamical system, numerical integration and function approximation provide the systematic approaches needed to alter the discrete forms.
In the context of numerical analysis, these types of transformations are referred to as time step refinement or mesh refinement.
Operations on the time step and integrator alter the computational graph and dictate the accuracy of the approximation.
Operations on basis function refinement give a way to project one tensor of trainable parameters onto a different tensor, with a potentially different shape.
Each of the following known transformations can be interpreted as the following operations on the computational graph or the tensor of trainable weights:
\begin{itemize}
	\item 
	{\bf Time-step refinement:} 
	Decrease or increase the number $N_t$ of time steps in the discrete representation. This shortens or deepens the computational graph.
	
	\item {\bf Integrator refinement:} Interchange the numerical integrator. This decreases or increases the approximation accuracy, and it alters the connectivity of the repeated graph modules.
	\item {\bf Weight refinement:}
	Decrease or increase the number $M$ of basis functions $\phi^\beta(t)$ in the approximation to $\theta(t)$.
	This decreases or increases the number of trainable parameters.
\end{itemize}

Time-step refinements and integrator refinements can be guided using numerical integration theory, and weight refinements can be guided using functional analysis. 

Because different numerical integrators should approximate the same continuous function, they can be freely interchanged or swapped.
In addition, the time step size $\Delta t$  that depends on $N_t$ can be changed in a way that minimally impacts model accuracy and smoothness.
In Sec. 3, these transformations were demonstrated on models trained for the nonlinear pendulum. 
In Sec. 5, these transformations will be demonstrated on an image classification problem. 
Finally, we can change the basis functions for $\theta(t)$ by projecting to a different basis, with either fewer functions (for a coarser approximation) or more functions (for a finer approximation).
The piecewise constant representation suggests a simple projection to twice as many intervals by ``splitting'' each interval into two domains and ``copying'' the value of each $\hat{\theta}^\beta$ into two trainable parameters, in a method similar to the proposed algorithm of \cite{chang2017multi}. Other projection and refinement strategies could be used.
Here, we consider uniform time-step sizes for the computational graphs and uniformly spaced intervals for basis functions; but here too other nonuniform distributions could be used.
For example, adaptive numerical integrators automatically adjust the time-step size, at the cost of additional computations and algorithmic logic.

\section{Empirical Properties of \CoolNets{}}
\label{sxn:refinenet-empirical}

In this section, we present empirical results illustrating the properties of \CoolNets{} for image classification problems. 
We show that \CoolNets{} can be compressed, i.e., the computational graph can be shortened, with little-to-no accuracy drop, which in turn leads to improved inference times (in Sec.~\ref{sxn:refinenet-empirical_classification}).
We also evaluate 
(i) weight refinement during the training process (in Sec.~\ref{sec:refinement_performance}), 
(ii) post-training time-step/depth refinement (in Sec.~\ref{sec:time_step_refinement}), and 
(iii) post-training integrator refinement (in Sec.~\ref{sec:integrator_refinement}).

\subsection{Application of \CoolNet{} to Image Classification}
\label{sxn:refinenet-empirical_classification}

We evaluate \CoolNets{} on both CIFAR-10 and CIFAR-100~\cite{krizhevsky2009learning}. 
Both datasets comprise 50K training images and 10K testing images, for 10 and 100 classes, respectively. 
(In Appendix~\ref{sec:app-additional-results}, we show similar results for Tiny ImageNet~\cite{wu2017tiny}.)
For this evaluation, we consider \CoolNets{} that replicate two canonical ResNet architectures (without bottleneck layers) that are known to perform well on the image classification tasks we consider: 
\begin{enumerate}
	\item \textbf{\CoolNet{}: } three blocks, with 16-32-64 channels, and up to 32-32-32 units (total depth 198); and
	
	\item \textbf{Wide-\CoolNet{}:} three blocks, with 64-128-256 channels, and up to 8-8-8 units (total depth 54).
	
\end{enumerate}
We evaluate \CoolNet{} on CIFAR-10, and we evaluate Wide-\CoolNet{} on CIFAR-100. 
For each problem, we train two versions: one version uses forward Euler units, and the other version uses RK4-Classic units.
The \CoolNet{}(Euler) networks are trained using a standard training methodology, by initializing the full network at $N_t=32$ for CIFAR-10 and $N_t=8$ for CIFAR-100. 
The \CoolNet{}(RK4-Classic) networks are trained using a \emph{refinement training method} (see Sec.~\ref{sec:refinement_performance}) to demonstrate the training capabilities provided by the \CoolNet{} framework.%
\footnote{We also trained \CoolNet{}(RK4-Classic) with a standard training methodology.  Except for training time, which we discuss below, the results were nearly identical to the results for \CoolNet{}(RK4-Classic) trained with the refinement training method.}
The models are implemented in PyTorch, and we use stochastic gradient decent with momentum~\cite{sutskever2013importance} for training.
Further, we use standard initialization schemes from the ResNet literature for the convolutional layers as well as for the linear and batch normalization layers~\cite{he2015delving}.

We use the notation $\mathrm{\CoolNet{}}[\mathtt{scheme},N_t]$ to represent the computational graph of the \CoolNet{} model, using integration scheme $\mathtt{scheme}$ and number of steps $N_t$.
Recall that $N_t$ is proportional to network depth as well as the computation time for a forward pass. 
Different schemes (forward Euler, RK-Classic, RK4-3/8, etc.) may also affect the effective depth and computation time.
The notation $\mathrm{\CoolNet{}}[\mathtt{scheme},N_t][\theta]$ defines plugging in a parameter function defined by its own weights $\theta(t;\theta^\beta)$.
We obtain the final discrete set of parameters $\theta^*$ by optimizing the cross-entropy $H$ for those two particular graph manifestations,
\begin{equation}
\theta^* = \arg \min_\theta \sum_{i\in train}H\left(y_i,\mathrm{\CoolNet{}}[\mathtt{scheme}^*,N_t^*][\theta](x_i)\right)  ,
\end{equation}
where the input and targets are denoted $x_i$ and $y_i$, respectively.
See Tab.~\ref{tab:cifar10_results} for a summary of our results.

\begin{table}[t] 
	\caption{Summary of classification accuracies for CIFAR-10 and CIFAR-100. The results show that \CoolNet{}(Euler) and Wide-\CoolNet{}(Euler) can be used as a 1-to-1 replacement for ResNet and Wide-ResNets. Further, \CoolNets{} outperform other related networks which are also nominally continuous-in-depth. 
	The results also show that the performance of \CoolNets{} is invariant both to changing number of time steps and to changing integrators. An asterisk (*) indicates that the identical trained weights are inserted from the parent model. Inference time reports the time to compute the error across the entire test set.}
	\label{tab:cifar10_results}
	\centering
	\begin{subtable}{0.95\textwidth}
		\caption{Results for CIFAR-10}
		\centering
		\scalebox{1.0}{
			\begin{tabular}{lcccccccccc}
				\toprule
				Model &  Units & Inference Time (s) & Params (M) & Accuracy &  Min/Max \\
				\midrule
				ResNet-200 (v2) &  32-32-32 & - & 3.19  & 93.84\% & 93.56\%/94.03\%\\
				ResNet-122-i \cite{chang2017multi} & 20-20-20 &-& 1.92 &  93.44\% & -\\
				Neural ODE \cite{zhang2019anodev2} & N/A &-& 0.45 & 67.94\% & 64.70\% / 70.06\%\\
				ANODEV2 \cite{zhang2019anodev2} & N/A & - &0.45 &  88.93\%& 88.65\% / 89.19\%\\
				Hamiltonian PDE \cite{ruthotto2019deep} & 3-3-3& - & 0.26 & 89.30\%& -\\
				
				\CoolNet{}(Euler)                 &  32-32-32 & 9.01 & 3.19 &   93.84\% & 93.55\% / 94.04\% \\
				\CoolNet{}(RK4-classic)           &  32-32-32 & 32.55 & 3.19 &  93.57\% & 93.40\% / 93.70\% \\
				$\quad \hookrightarrow$(Euler)   & 32-32-32  & 8.93 & (*) &  93.55\% & -  \\
				$\quad \hookrightarrow$(RK4-3/8) & 11-11-11  & 11.06 &(*) &   93.44\% & - \\
				$\quad \hookrightarrow$(RK4-3/8) & 6-6-6     & 6.25 & (*) &   92.28\% & - \\
				
				\bottomrule
		\end{tabular}}
		\label{tab:table1_a}
	\end{subtable}\vspace{+0.4cm}
	
	\begin{subtable}{0.95\textwidth}
		\caption{Results for CIFAR-100}
		\centering	
		\scalebox{1.0}{
			\begin{tabular}{lcccccccccc}
\toprule
Model &       Units & Inference Time (s) & Params (M) & Accuracy &  Min/Max \\
\midrule
Wide-ResNet-50 ($4\times$) & 8-8-8 & - & 13.61 & 79.04\% & 78.67\%/79.35\% \\ 
Hamiltonian PDE \cite{ruthotto2019deep} & 3-3-3 & - & 0.36 & 64.90\%& -\\
Second-order PDE \cite{ruthotto2019deep} & 3-3-3 &- & 0.65 & 65.40\%& -\\

Wide-\CoolNet{}(Euler)            & 8-8-8 & 4.20 & 13.58 & 		78.80\% & 78.20\% / 79.65\%  \\
Wide-\CoolNet{}(RK4-Classic)      & 8-8-8 & 14.56 & 13.58 &  77.99\% & 77.68\% / 78.12\% \\
$\quad \hookrightarrow$(RK4-3/8) & 4-4-4 & 7.74 & (*) & 76.19\% &  - \\
\bottomrule
		\end{tabular}}	
		\label{tab:table1_b}
	\end{subtable}
\end{table}

In Tab.~\ref{tab:cifar10_results}, \CoolNet{}(Euler) refers to the model trained using $\mathrm{\CoolNet{}}[\mathtt{Euler},N_t=32]$, with weights $M=32$, whose computational graph is equivalent to a ResNet with 32-33-33 units in each block. 
(The count of ResNet units is $N_t+1=33$ in the second and third blocks to account for the stitch down-sampling connections in \CoolNet{}.) 
Wide-\CoolNet{}(Euler) corresponds to Wide-$\mathrm{\CoolNet{}}[\mathtt{Euler},N_t=8]$ with $M=8$, which is equivalent to a Wide-ResNet 8-9-9.
Likewise, \CoolNet{}(RK4-Classic) refers to the model which {\em ended} training as a \CoolNet{}[RK4-Classic, $N_t=32$], with weights $M=32$; and Wide-\CoolNet{}(RK-Classic) corresponds to  \CoolNet{}[RK4-Classic, $N_t=8$] with weights $M=8$.
For these models, however, we used the \emph{refinement training method} (see Sec.~\ref{sec:refinement_performance}) to learn the weights $\theta(t)$.

From Table~\ref{tab:table1_a} and Table~\ref{tab:table1_b}, we see several trends in classification accuracies for CIFAR-10 and CIFAR-100, respectively. 
First, we see that the \CoolNet{}(Euler) models yield similar generalization errors as do ResNets.
We establish this baseline by considering a standard ResNet-200 and Wide-ResNet-58 (not the \CoolNet{}-based implementation, but a standard implementation trained on our machine with tuning parameters known from the literature.)
Next, we see that the generalization errors of \CoolNet{}(RK4-Classic) are (slightly) worse. 
This is not surprising, since the RK4-Classic scheme introduces a different \emph{inductive bias} into the training process.
This bias in turn also has benefits such as manifestation invariance and improved stability with respect to input perturbations, as we discuss below.
Finally, we see that, in contrast to other related models, including ANODEV2~\cite{zhang2019anodev2} and Hamiltonian PDE~\cite{ruthotto2019deep}, our \CoolNets{} can train deeper architectures that are on par with standard ResNet implementations. 

In the next two subsections, we go into more detail on these results, with an emphasis on how \CoolNet{}'s continuous-in-depth structure can enable improved training and improved inference.

\subsection{Improved Training: Weight Refinement During the Training Process}
\label{sec:refinement_performance}

Here, we describe how the continuous-in-depth property of \CoolNet{} can be used to develop an iterative \emph{refinement training method}.
(The versions of \CoolNet{}(RK4-Classic) reported in Tab.~\ref{tab:cifar10_results} were trained with this method.)
This method implements a form of \emph{weight refinement} during the training process, by first initializing with a shallow NN and then iteratively deepening the model as training progresses.
It can be used to improve the approximately $4\times$ extra cost associated with a na\"{\i}ve use of the RK4-Classic method in place of the forward Euler method.

In more detail, each OdeBlock is initialized with $N_t=1$; and then, at preset epochs, we perform a refinement step in which we simultaneously double $N_t$ (halving $\Delta t$) and double $M$ (splitting the basis functions).%
\footnote{This refinement process is similar to one downward cycle of a multigrid method used for solving partial differential equations~\cite{ASTRAKHANTSEV1971171}.}
Importantly, due to the continuous-in-depth property of \CoolNets{}, the effective meaning of the network stays unchanged during this refinement, by the implied $\Delta t$ scaling. 
(In particular, this means that we are able to tune hyperparameters on coarse networks, and then train deep networks with these hyperparameters unchanged.)
Refinements can be executed at various stages of the training process.
Here, we consider five refinement steps for CIFAR-10 and three refinement steps for CIFAR-100.
The deeper network was refined at epochs [20, 40, 60, 70, and 80],
and the shallower network was refined at epochs [20, 50, and 80].

\begin{figure}[!t]
    \centering
    \includegraphics[width=0.9\linewidth]{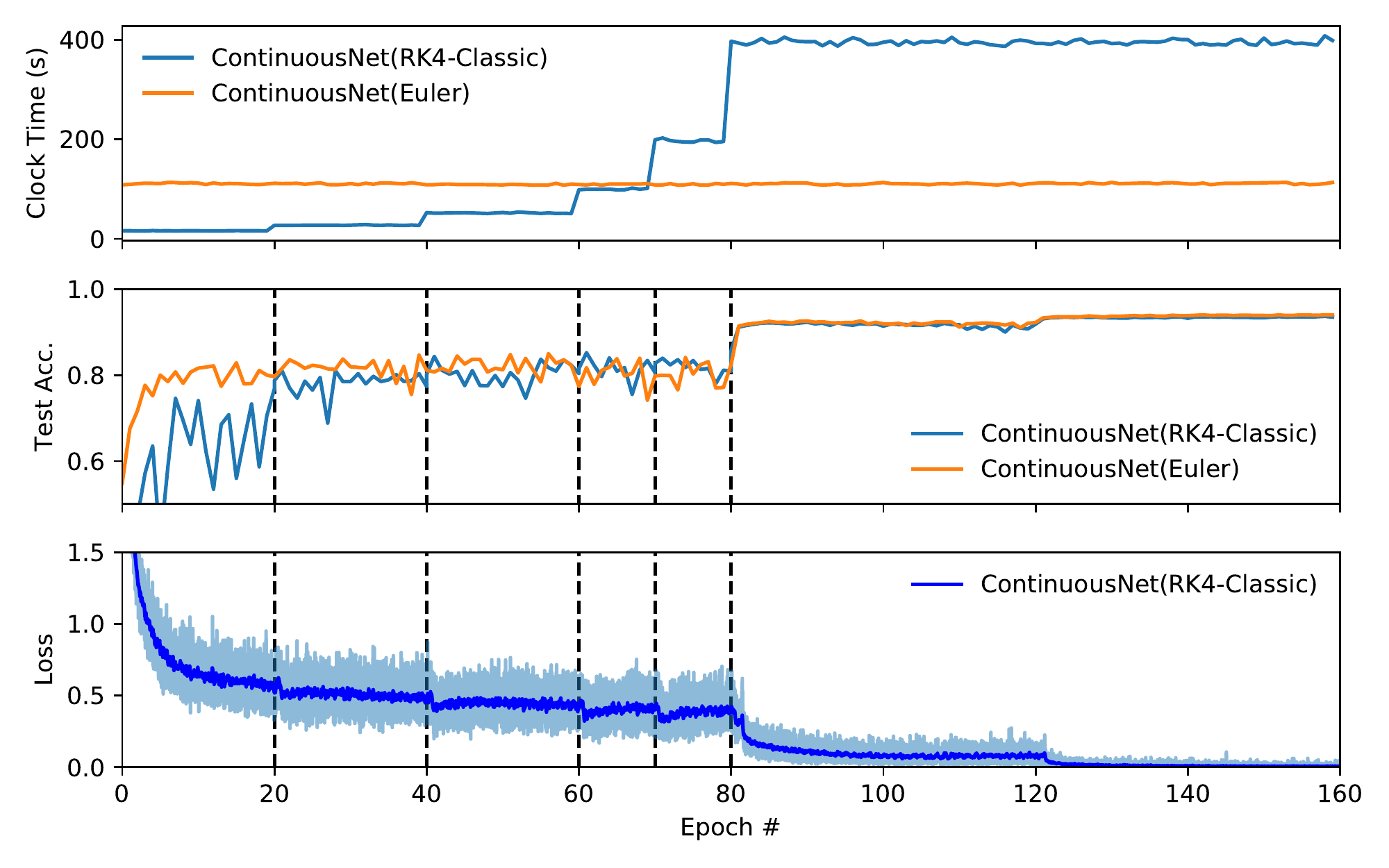}
    \caption{Progression of training behavior showing (top) clock time per epoch to compute the gradient update at each epoch, (middle) test accuracy after each epoch, and (bottom) loss function value after each epoch (light blue) with a running-average (dark blue). The \CoolNet{}(RK4-Classic) network was trained with the refinement algorithm, with refinement steps denoted by vertical lines, which impacts its training progression and run-time efficiency.
    The run-times and test accuracy of the \CoolNet{}(Euler) network (trained without the refinement algorithm) is included as a point of comparison.
    At each refinement, the weight parameters in each OdeBlock were split (which doubled the number of parameters), and the number of time steps $N_t$ was doubled. 
    Note that the test accuracy and loss improve immediately after refinement.
    The run-time efficiency of the \CoolNet{}(RK4-Classic) is dramatically improved using the algorithm, even though each individual epoch is ultimately more expensive at the end phase of training when the network is at its final parameter count and graph depth.
	}
    \label{fig:loss_functions}
\end{figure}

To illustrate this on CIFAR-10, the middle and bottom panels of Fig.~\ref{fig:loss_functions} show the test accuracy and loss during training for the \CoolNet{}(Euler) network (trained without the refinement algorithm) and for the \CoolNet{}(RK4-classic) network (trained with the refinement algorithm). 
When refined at epochs [20, 40, 60, 70, and 80], \CoolNet{}(RK4-Classic) yielded a final size of $N_t=M=32$ for each OdeBlock and the parameter counts at during this process are 0.2M, 0.3M, 0.5M, 0.9M, 1.6M and 3.2M.
At $N_t=M=32$, the total depth would correspond to a ResNet-198 if it were evaluated with a forward Euler scheme manifesation.
No detrimental effects occur during refinement, and the model accuracy and loss improve slightly after refinement, as desired. 
Thus, no alterations are required from a learning rate decay schedule that would be used for a deeper network.
This contrasts sharply with the refinements that are implemented for models that are not continuous-in-depth~\cite{chang2017multi}. 

To avoid the possibility of becoming trapped in a minimum corresponding to a shallower network, the epochs at which refinement occurred were chosen to be just before the learning rate was decayed. 
Beyond that requirement, we did not perform hyperparameter tuning to determine an optimal refinement schedule.
How to optimize training speed, e.g., by finding the latest epoch at refinements could occur without affecting model accuracy, is an interesting direction for future~work. 

The top panel of Fig.~\ref{fig:loss_functions} shows the wall-clock run time for the training step of each epoch, neglecting test accuracy evaluation.
(In particular, time-per-epoch is constant across training for Euler since we used the standard training methodology, and time-per-epoch is less for earlier epochs than for later epochs for RK4-Classic since we used the refinement training method.)
On a single nVidia Tesla V100, each of the first 20 epochs for \CoolNet{}(RK4-Classic) took $\sim$16s (when $N_t=1$) and each of the last 80 epochs took $\sim$394s (when $N_t=32$).
Since the earlier epochs work with a smaller model, the earlier epochs are significantly faster than epochs during the last half of training.
As a point of comparison, recall that using RK4-Classic (the standard training methodology) is significantly slower for a given depth than using Euler, since it requires four times as many evaluations of $\mathcal{R}$ as forward Euler does. 
The \CoolNet{}(Euler) network required $\sim$113s per epoch, with $N_t=32$ and $\sim$295m overall. %
Hence, our refinement training method for \CoolNet{}(RK4-Classic) accelerated the overall training process, i.e., the entire training time was reduced from $\sim$1053m (when training with fixed $N_t=32$, i.e., no refinement training method) to $\sim$608m (with the refinement training method).
While still slower than \CoolNet{}(Euler), our refinement training method for \CoolNet{}(RK4-Classic) takes only roughly twice as long, as opposed to roughly four times as long, had we not used the refinement training method.

Note that we used the na\"{\i}ve graph back-propagation algorithm. 
Adjoint-based differentiation is possible with \CoolNets{}, as it theoretically has asymptotically lower memory efficiency.
However, we observed that the existing Python implementation was slower than using back-propagation.
We also observed that the adjoint gradients were sensitive to the number of time steps: for the network to converge to a high accuracy, the time steps needed to be one third the width of the basis function intervals, i.e., $N_t\ge 3M$.
A similar observation was discussed in \cite{gholami2019anode}.

 \begin{figure}[!t]
 \centering
  \includegraphics[width=0.45\textwidth]{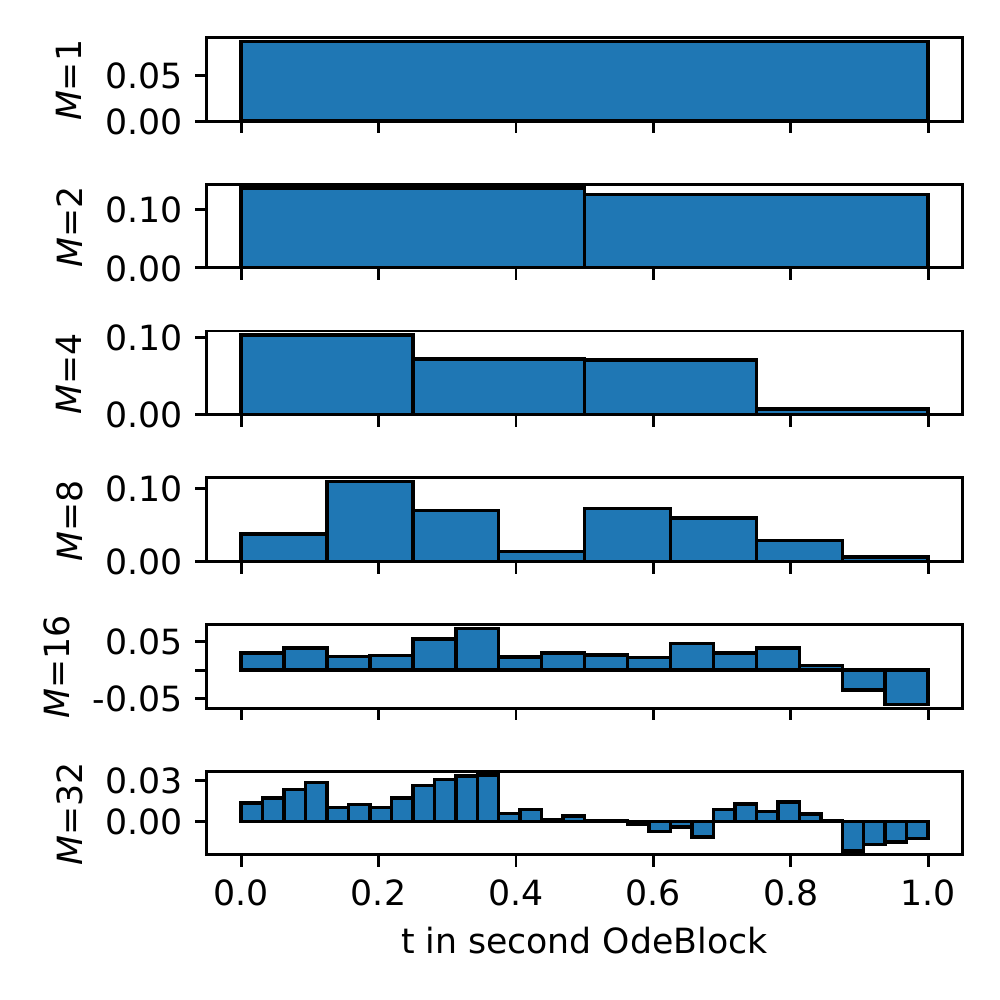}%
  \includegraphics[width=0.45\textwidth]{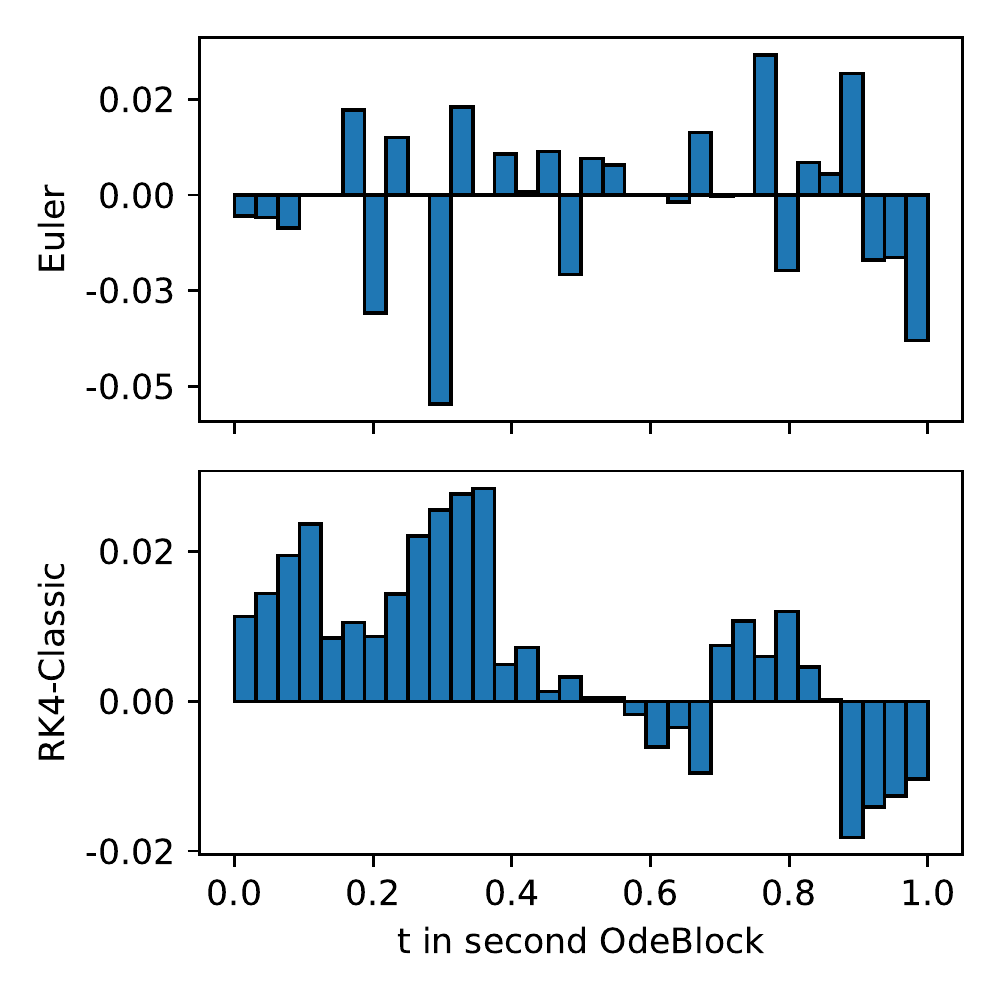}
 \caption{Illustration of how weights of a \CoolNet{} are refined during training. 
 Weights are piecewise constant functions, defined over $t\in[0,1]$, and divided into a varying number of $M=N_t$ segments. 
 On the left, \CoolNet{}(RK4-Classic) was refined to finer discretizations during training, starting with only one weight at initialization (top), which iteratively splits until there are 32 weights (bottom). On the right, we illustrate (with a zoom-in) the final weights for \CoolNet{}(Euler) and \CoolNet{}(RK4-classic). 
 Compared to \CoolNet{}(Euler), 
 transitions between layers are smoother when \CoolNet{}(RK4-classic) is trained with refinement. 
 }
 \label{fig:basis_functions}
 \end{figure}

Fig.~\ref{fig:basis_functions} shows the evolution of the weights of the CIFAR-10 \CoolNet{}(RK4-Classic) instance at different stages of weight refinement during training.
The intermediate weights are not converged, but they are snapshots at the refinement epochs (20, 40, 60, 70, and 80).
This illustrates that the weights learn a smoother function when using the combination of \CoolNet{}(RK4-Classic) and the iterative refinement training method.
In contrast, the \CoolNet{}(Euler) model shows that the weights are much less smooth across time intervals.
This is consistent with~\cite{rothauge2019residual}.

\subsection{Improved Inference: Exploiting Manifestion Invariance}

Here, we describe how the continuous-in-depth property of \CoolNet{}---and in particular how continuous-in-depth models should be robust to the manifestation of the particular time-step/depth (\emph{time-step refinement}) and the particular discrete numerical integrator (\emph{integrator refinement})---can be used to improve inference in the post-training stage.

In more detail, to understand how \CoolNets{} exhibit the properties of a continuous dynamical system in a ``real'' ML problem, and not just for learning the continuous dynamical system in Sec.~\ref{sec:overview}, we study the behavior of trained \CoolNets{} while we alter (i) the time step and/or (ii) the ODE integrator scheme.
To do so, we use a  procedure that is similar to the convergence analysis that was shown in Fig.~\ref{fig:pendulum_theta_convergence}.
However, here we don't have access to a ground truth function since we do not know a governing equations that describes the dynamics.%
\footnote{Indeed, using a \CoolNets{} model amounts to assuming that such a continuous structure exists---it is a modeling assumption.}
Instead, we examine whether the test error smoothly varies as we decrease $\Delta t=N_t/T  $, i.e., as we increase $N_t$, the number of layers. 
We also examine how the test error varies if we use a NN constructed from one ODE integrator for training and a different ODE integrator for inference.

\begin{figure}[!t]
	\centering
	\begin{subfigure}[t]{0.9\textwidth}
		\centering
		\begin{overpic}[width=1\textwidth]{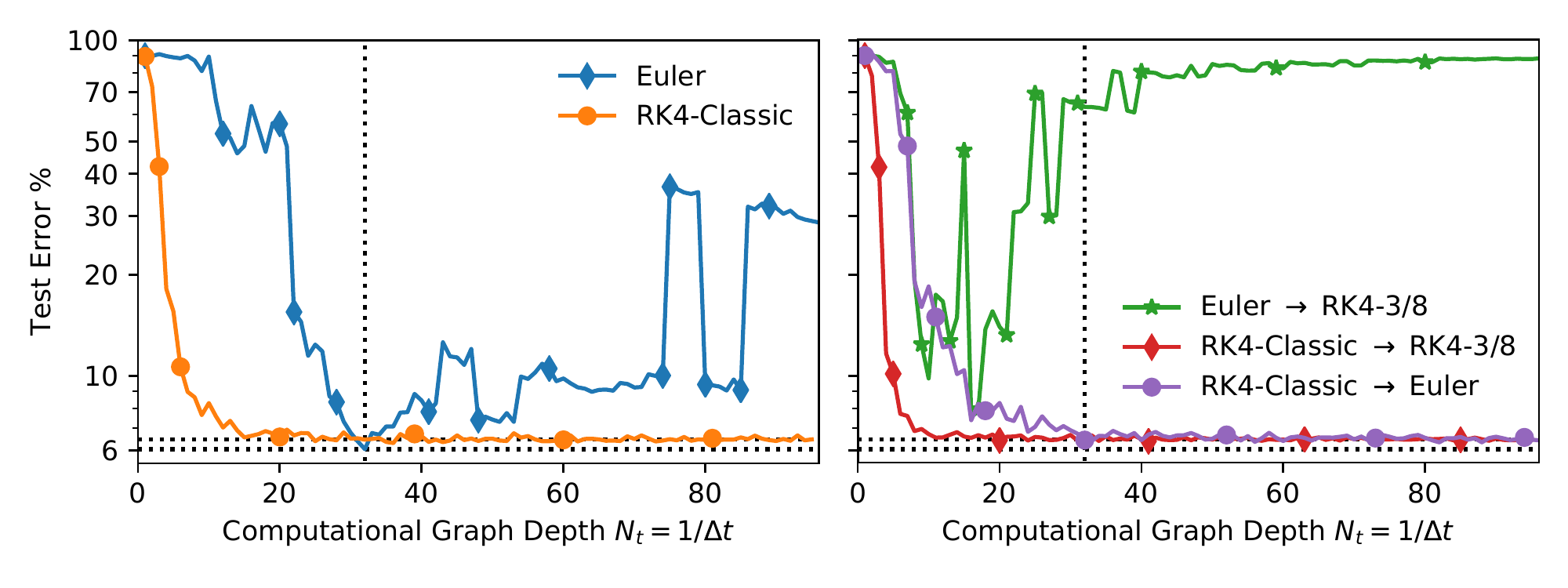}
		\end{overpic}\vspace{-0.4cm}	
		\caption{CIFAR-10}
	\end{subfigure}\vspace{-0.0cm}
	
	\begin{subfigure}[t]{0.9\textwidth}
		\centering
		\begin{overpic}[width=1\textwidth]{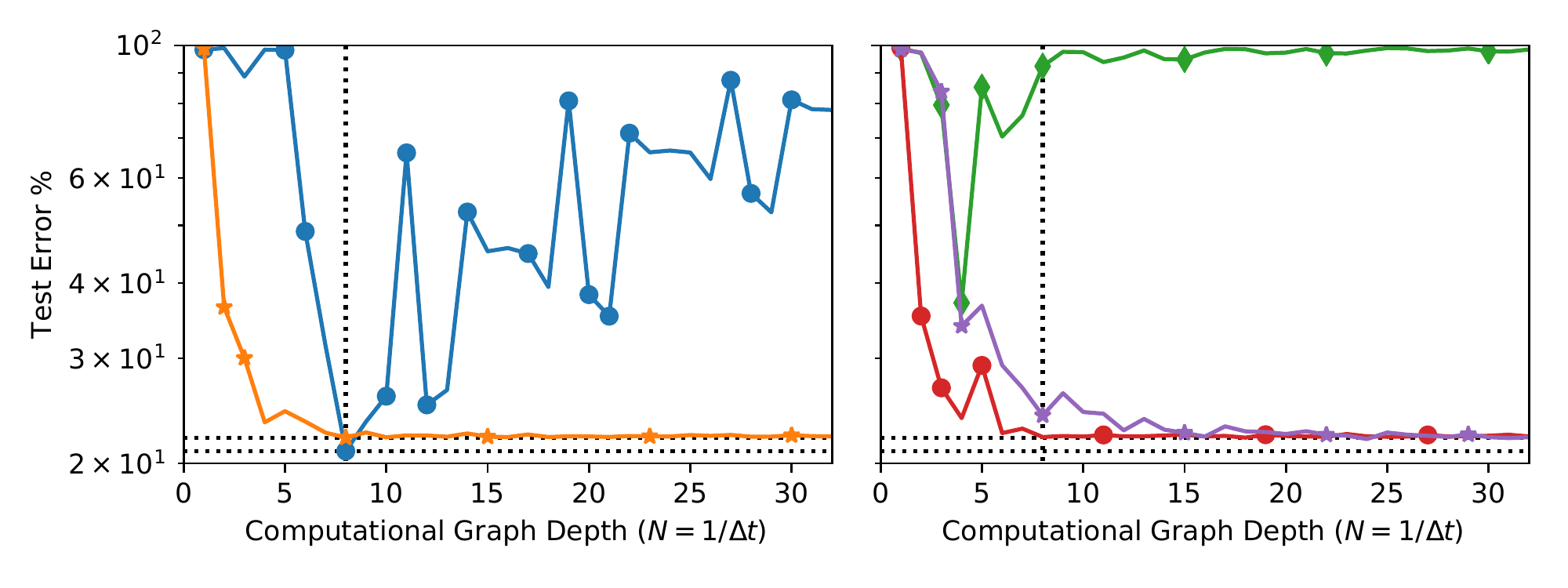} 
		\end{overpic}\vspace{-0.4cm}			
		\caption{CIFAR-100}
	\end{subfigure}      
	
	\caption{
		\CoolNets{} are more robust when trained with RK4-Classic than forward Euler, since higher-order architectures introduce an inductive bias that smooths across changing depth. Each data point represents a unique graph manifestation resulting from a choice for $\Delta t$ and/or the integrator. The vertical dotted lines indicate $N_t$, and horizontal lines indicate test error on the original architecture. 
		(Left)~ResNet-equivalent models have behavior that changes dramatically after any minor alteration in depth, whereas \CoolNet{}(RK4-Classic) converges to a smooth dynamical system as $\Delta t\to 0$ and thus have a much more robust behavior as depth is varied.
		(Right)~Properly trained \CoolNets{} remain accurate after interchanging the integrator, which yields different computation graphs (i.e., NN architectures).
		This can either  decrease model accuracy and performance (e.g., RK4-classic $\to$ forward Euler), or it can improve model accuracy and performance (e.g., RK4-classic $\to$ RK4-3/8).
		}
	\label{fig:cifar10_manifestation}
\end{figure}

Altering the time step or exchanging the ODE integrator scheme both generate a new computational graph, i.e., for each combination $(scheme,N_t)$ that we consider, we obtain a new graph that reuses the pretrained weights $\theta^*$.%
\footnote{To reiterate, the usual perspective would be that each of these networks is a different discrete NN model; but, from the continuous-in-depth perspective, each of these networks is a different discrete approximation to the same continuous-in-depth model.}
Then, each of these new graphs is evaluated on the test set:
\begin{equation}
E_{test}(N_t,scheme) = 1-\frac{1}{|test|}\sum_{i\in test}\left(\mathrm{\CoolNet{}}[scheme,N_t][\theta^*](x_i)=y_i\right)  .
\end{equation}
Fig.~\ref{fig:cifar10_manifestation} shows the test errors obtained for the different computational graphs for CIFAR-10 (top) and CIFAR-100 (bottom). 
We see that higher-order integrators introduce an inductive bias, which in turn leads to models that are \emph{much} smoother, and thus that represent in a meaningful sense a continuous dynamical systems.
This is because the accuracy of \CoolNets{} trained in RK4-classic does not degrade when more (or fewer, up to a point) time steps are added, or when the integration scheme is exchanged. 
This is consistent with our analysis of the pendulum in Sec.~\ref{sec:overview}, and it means that the \CoolNets{} model is not over-fitting to the number of layers.

\subsubsection{Time-step Refinement: Forward Euler versus Smooth \CoolNets{}}
\label{sec:time_step_refinement}

In the left panels of Fig.~\ref{fig:cifar10_manifestation}, we show that a trained \CoolNet{} model can be modified by changing the number of time steps $N_t$ (i.e., units) in each OdeBlock.
We compare the graphs that use either the weights $\theta^*$ trained by \CoolNet{}(Euler) or those trained by \CoolNet{}(RK4-Classic). 
The vertical dotted lines indicate the $N_t$ values that were used for training.
Observe that this plot has a similar structure to Fig.~\ref{fig:pendulum_theta_convergence}, but here we don't use log-log plots (and we don't have a ground truth to compare against).
The horizontal axis plots number of steps $N_t$, instead of $\Delta t = 1/N_t$, which corresponds to network depth and to computation time, respectively, such that going to the right increases cost and should increase accuracy.
(These plots are \emph{not} smooth because the piecewise constant functions for $\theta(t)$ are not Lipschitz continuous along boundaries, and thus smooth convergence is not theoretically guaranteed.)

Observe that the error curves for the graphs that use the weights $\theta^*$ trained by \CoolNet{}(Euler) have  significant ``dips.'' 
In particular, they have a sharp minimum at the time-step/depth at which the model was trained, which then degrades substantially for other values of the time-step/depth. 
These dips are analogous to the sharp dip we observed (for Euler) in Fig.~\ref{fig:pendulum_theta_convergence_a}. 
(Here, we also observe additional dips, possibly due to harmonics of the architecture.) 
That is, increasing or decreasing the number $N_t$ of layers, even slightly, drastically changes performance.
This indicates that the model, and in particular the ``temporal discretization,'' overfits the data, and thus the model does \emph{not} represent a smooth dynamical system in a meaningful sense.
When using Euler, this model is a good discrete model, but it is not a good continuous model.

This behavior contrasts sharply with the \CoolNet{}(RK4-classic) model. 
(Note that we do \emph{not} retrain the weights here.)
The errors for the \CoolNet{}(RK4-classic) model varies smoothly and remains fairly accurate as a function of the depth/time-step of the computational graph. 
This smoothness is analogous to the smoothness we observed (for Midpoint and RK4) in Fig.~\ref{fig:pendulum_theta_convergence_a}.
When using RK4-classic, this \CoolNet{}(RK4-classic) model is a good discrete model, and it is also a good continuous model.

\subsubsection{Integrator Refinement: \CoolNet{} Invariance Properties}
\label{sec:integrator_refinement}

In the right panels of Fig.~\ref{fig:cifar10_manifestation}, we show that a trained \CoolNet{} can be modified by exchanging the integrator in each OdeBlock---at the inference step, i.e., without any re-training. 
Again, we compare graphs that use either the weights $\theta^*$ trained by \CoolNet{}(Euler) or the weights trained by \CoolNet{}(RK4-Classic). 
We observe that the error curves that corresponds to the graphs using weights trained by \CoolNet{}(RK4-Classic) do not degrade and do not fluctuate, as $N_t$ changes. 
In stark contrast, the graphs that uses the weights trained by \CoolNet{}(Euler) shows a catastrophic performance when the integrator is exchanged. 
These results are analogous to the interchangeability results presented in
Fig.~\ref{fig:pendulum_theta_convergence_b}.

In Table~\ref{tab:table1_a}, we include examples for changing the integrator of the \CoolNet{}(RK4-classic) instance with max accuracy (93.70\%). 
The learned weights are just as accurate when the integrator is interchanged to forward Euler; and, in addition, the model is {\em also as good as a ResNet.} 
(Importantly, the converse is not true.)
Further, we can interchange integrators and compress the graph depth from $N_t=$32-32-32 down to $N_t=$11-11-11, while retaining an accuracy within the minimum and maximum of the repeated training samples of the unmodified network. 
The graph can be compressed even further down to $N_t=$6-6-6, yielding a slight drop in accuracy, 92.28\%, despite the model being significantly shorter. 
A similar performance is obtained by Wide-\CoolNets{} on CIFAR-100.
In turn, the inference time (for classifying the test data) reduces significantly.
We stress that the weights were \emph{not} retrained or modified during this architecture~swap.

\section{Conclusion}
\label{sxn:conc}

Based on recent work viewing ResNets as dynamical systems, and in particular as discrete numerical integrators of continuous dynamical systems, we asked to what extent does a model exhibit the expected properties of a numerical integrator, beyond simply having a similar syntactic algebraic form.
By applying convergence tests from numerical analysis, we were able to demonstrate that ResNets do not in fact learn to represent continuous dynamical systems in any meaningful sense, either for prototypical physical time series problems or for prototypical computer vision problems. 
However, using a clarified understanding of the relationship between numerical integrators and NN models that are trained on data, we developed \CoolNets{} as a meaningfully continuous-in-depth generalization of ResNets.  
Our main results are the~following:
\begin{itemize}
		\item 
		We show that despite their algebraic similarity, ResNets (as well as forward-Euler-based NN-based models for ODEs) do not correspond to forward Euler. 
		ODE-based models require higher-order integrators to avoid overfitting to the time step size in order to appropriately represent a continuous dynamical system.
		\item 
		We establish a direct and principled intersection between ResNets and  continuous-in-depth NN-based ODEs. 
		The resulting \CoolNets{} outperform existing continuous-in-depth models with respect to growing very deep networks and achieving state-of-the-art accuracy.
		\item 
		We demonstrate that higher-order integrators (e.g., Runge-Kutta methods) introduce an inductive bias, allowing \CoolNets{} to balance accuracy and robustness. In contrast, other state-of-the art ResNets are good discrete models, but they are not good continuous models. 
		
		\item 
		We show that when a \CoolNet{} appropriately represents a continuous dynamical system, it exhibits {\em manifestation invariance}. 
		Manifestation invariance simplifies and improves the robustness of iterative-refinement training methods; and it also allows a trained \CoolNet{} to be seamlessly compressed to smaller architectures at inference time, without the need to revisit the data.
\end{itemize}

As part of our approach, we have introduced the property of manifestation invariance, whereby a given model description can be transformed into various  discrete models.
\CoolNet{} models achieve this by the application of existing tools from numerical analysis, such as using ``mesh refinement'' to grow in depth during training and shorten computation during inference.
Importantly, these operations do not require revisiting the data to create new discrete network descriptions of the learned continuous operation.
Our empirical results validated this property and showed that the performance
(based on standard ML generalization accuracy on a test set) of \CoolNets{}
are comparable to that of traditional ResNets.

Based on our results, there are several interesting directions for future work, including the following immediate plans:

\begin{itemize}

    \item 
    Basis functions: 
    In this work, we only considered piecewise constant basis functions, which allowed us to recover the same behavior as ResNets.
    However, there exist a large number of other interesting basis functions that have potential advantages. 
    Future work should explore potential candidate basis functions as well as their tradeoffs with respect to convergence and accuracy. 

    \item 
    Integrator schemes: 
    In this work, we considered one-step integrators from the explicit Runge-Kutta family. 
    However, there exist a vast number of other interesting integrators, including implicit solvers and adaptive time step solvers. 
    Future work should explore tradeoffs when using more advanced integrator schemes. 

    \item 
    Compression: 
    In this work, we demonstrated that \CoolNets{} are highly compressible. 
    Since \CoolNets{} are more robust with respect to parameter perturbations, we expect that \CoolNets{} trained with our refinement training methodology should be well-suited to compression and other quantization strategies. 
    Future work should explore how this idea can be exploited to design efficient networks for edge devices.

    \item 
    Transfer learning: 
    In this work, we demonstrated that \CoolNets{} can be easily extended into a deeper architecture during and after training. 
    This feature might be interesting for transfer learning, where one typically aims to fine-tune the last few layers or ResNet units for a new target problem. 
    Future work should explore the transfer learning properties of extended \CoolNets{} and their refinement capabilities.
    
    \item
    Hyperparameter optimization: The continuous-in-depth framework yields consistent training behavior among \CoolNets{} at different refinement depths and hyperparameter choices. 
    This may aid in hyperparameter optimization,
    wherein an important real-world cost is the length of time required to perform the full hyperparameter search and then train the final model.
    Future work should explore improvements to hyperparameter search that are possible within the continuous-in-depth framework.
    
\end{itemize}

Of course, it is not necessary that a ML problem be solved by a continuous dynamical system.
ResNets do solve certain tasks well.
However, they do not correspond to a dynamical system unless the additional measures proposed for \CoolNet{} are imposed.
Essentially, this gives us an additional restriction on the class of models we consider, and it leads to many benefits for both training and inference.
\CoolNets{} based on high-order integrators introduce an inductive bias that stabilizes the networks, leading to more robust models, and allowing them to grow into deeper and more accurate models than other continuous-in-depth models, such as the Neural ODE-based approaches~\cite{chen2018neural,zhang2019anodev2}.

\section*{Acknowledgments}
We are grateful for the generous support from Amazon AWS and Google Cloud.
A. F. Queiruga would like to acknowledge the US Department of Energy for providing partial support of this work.
D. Taylor would like to thank the Simons Foundation for providing partial support of this work.
N. B. Erichson and M. W. Mahoney would like to acknowledge the UC Berkeley CLTC, ARO, IARPA, NSF, and ONR for providing partial support of this work.
Our conclusions do not necessarily reflect the position or the policy of our sponsors, and no official endorsement should be inferred.

{\normalsize
\bibliographystyle{plainnat} 
\bibliography{shallow} 
}

\input{appendix.tex}

\end{document}

%% file: appendix.tex
\appendix

\clearpage
\section{Additional Details on Continuous Network Elements}
\label{sec:app-additional-details}

Here, we provide additional details on the architecture required to combine continuous and discrete network elements. 
These considerations are not fundamental to the definition of continuous-in-depth networks, and they represent migrations of existing concepts to the continuous-in-depth framework. 
As with designing ResNets, these network ``features'' are critical to maximizing performance.

\subsection{Stitching Between Continuous OdeBlocks}

Modern ResNets for image classification perform reshaping steps at certain points in the network.
For example, for CIFAR-10 the dimensions of $x$ are down sampled from $32\times32$ to $16\times16$ to $8\times8$, in between blocks of pure residual units. 
This breaks the continuity of the ODE chain because the number of time-dependent variables changes, and thus this requires special consideration.

\CoolNet{}s replicate the reshaping operations of standard ResNet architectures by borrowing one discrete downsampling residual unit.
The three OdeBlocks are stitched together, as illustrated in Fig.~\ref{fig:odenet}, using a discrete ``stitch''~unit:
\begin{equation}
\label{eq:stitch}
    x_{out} = \mathtt{downsample}(x_{in};\mathtt{in}=32,\mathtt{out}=16) + \epsilon \mathcal{R}(x_{in};\theta_k; \, \mathtt{in}=32,\mathtt{out}=16)
\end{equation}
for the first block. 
This is identical to the first unit in a residual block in discrete ResNets.
To scale the magnitude of the $\mathcal{R}$ within the stitch to that of the $\mathcal{R}$ within the ODEBlock, we set $\epsilon_{stitch}=\epsilon_{ode} T/\Delta t$. 
For image classification, the two stitch segments described by Eq.~\eqref{eq:stitch} are used between the three continuous blocks, as illustrated by Fig.~\ref{fig:odenet}.

\subsection{Time Dependent Skip Initialization Factors}

Following a variation proposed for discrete ResNets by~\cite{de2020batch}, an architecture variation was used wherein ``skip initialization'' terms are included. 
These terms are used in conjunction with BatchNorm elements.
In this variation, the equation for $\dot{x}$ has a ``skip'' term that is a scalar multiplier,
\begin{equation}
    \mathcal{R}(x,t)=s(t) \left( \mathtt{conv}\left(\sigma(\dots) \right)\right).
\end{equation}
The weights of the convolutions are initialized in the usual manner, and the skip term $s$ is initialized to zero, making the OdeBlock an identity at the first iteration. 
The function $s(t)$ is treated as a component of $\theta(t)$. 
The function $s(t)$ is parameterized in the way described in the main text, with its own coefficients $s(t)=\sum s^\beta \phi^\beta (t)$. 
The same basis functions are used for the other weights, as described in the main text.
Each OdeBlock has its own $s(t)$ function.
This time-dependent form also algebraically matches exactly a discrete ResNet, using this residual module variation in the forward Euler limiting-case of a \CoolNet{}.

\subsection{Time Dependent Batch Normalization}

Batch normalization does not fit easily into the continuous-in-depth viewpoint. 
To implement this useful procedure, we define a ``continuous-time batch normalization'' as being a normalization that is sliced up along the same piecewise-constant intervals used for the weight bases.
That is, each interval has its own BatchNorm operator.
The data structure representing a continuous-time BatchNorm operation consists of a list of discrete BatchNorm objects and then serves to index into this list during evaluation, based on the time:
\begin{equation}
    \mathtt{bn}(t,x) = \begin{cases}
        \mathtt{bn}^1(x) & ;\,\, t\in [0, \Delta t) \\
        \mathtt{bn}^2(x) & ;\,\, t\in [\Delta t, 2 \Delta t) \\
        ... \\
        \mathtt{bn}^N(x) & ;\,\, t\in [t-\Delta t,T].
    \end{cases}
\end{equation}
Each of these objects has its own affine parameters and keeps statistics independently. 
When the time-dependent batch normalization module is refined, the number of indicator functions in the basis doubles, each of the independent $\mathtt{bn}(x)$ modules is copied, and the two copies are assigned to corresponding indices in the new set of indicator functions.

%
%

\section{Additional Results}
\label{sec:app-additional-results}

\begin{figure}[!t]
    \centering
    \includegraphics[width=0.9\textwidth]{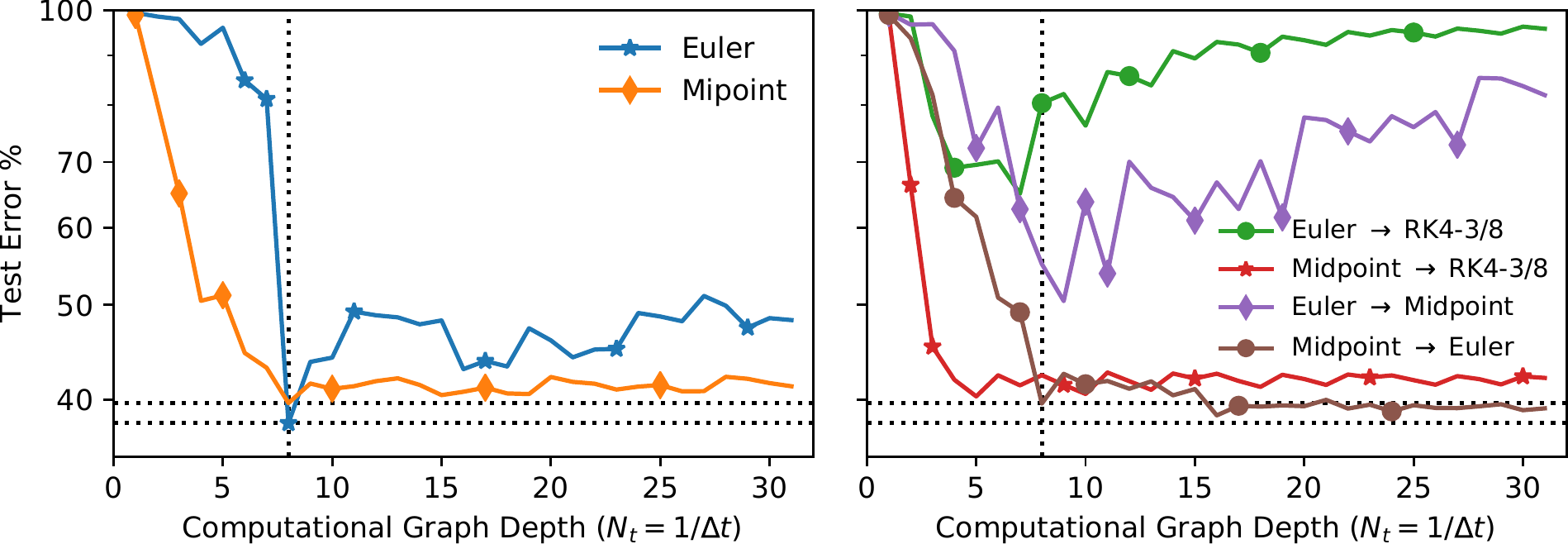}
    \caption{\CoolNet{} trained on Tiny-Imagenet using Euler and Midpoint with the wide architecture used for CIFAR-100.  Test error as a function of depth (left); and behavior under interchanging the numerical integrator (right).  The bottom horizontal line is the baseline accuracy of \CoolNet{}(Euler), and the upper horizontal line is that of \CoolNet{}(Midpoint). While this architecture was not designed to perform well on Tiny-ImageNet, it is evident that the same observations are made, as compared to the CIFAR models in the main text. The Euler-based model overfits to the depth/time, and it cannot be interpreted as a continuous dynamical system, while the \CoolNet{}(Midpoint) model exhibits the similar accuracy after altering the timestep size and the integration scheme used to generate its graph.
    }
    \label{fig:tiny}
\end{figure}

\begin{figure}[!t] 
	\centering
	\begin{subfigure}[t]{0.9\textwidth}
		\centering
		\begin{overpic}[width=1\textwidth]{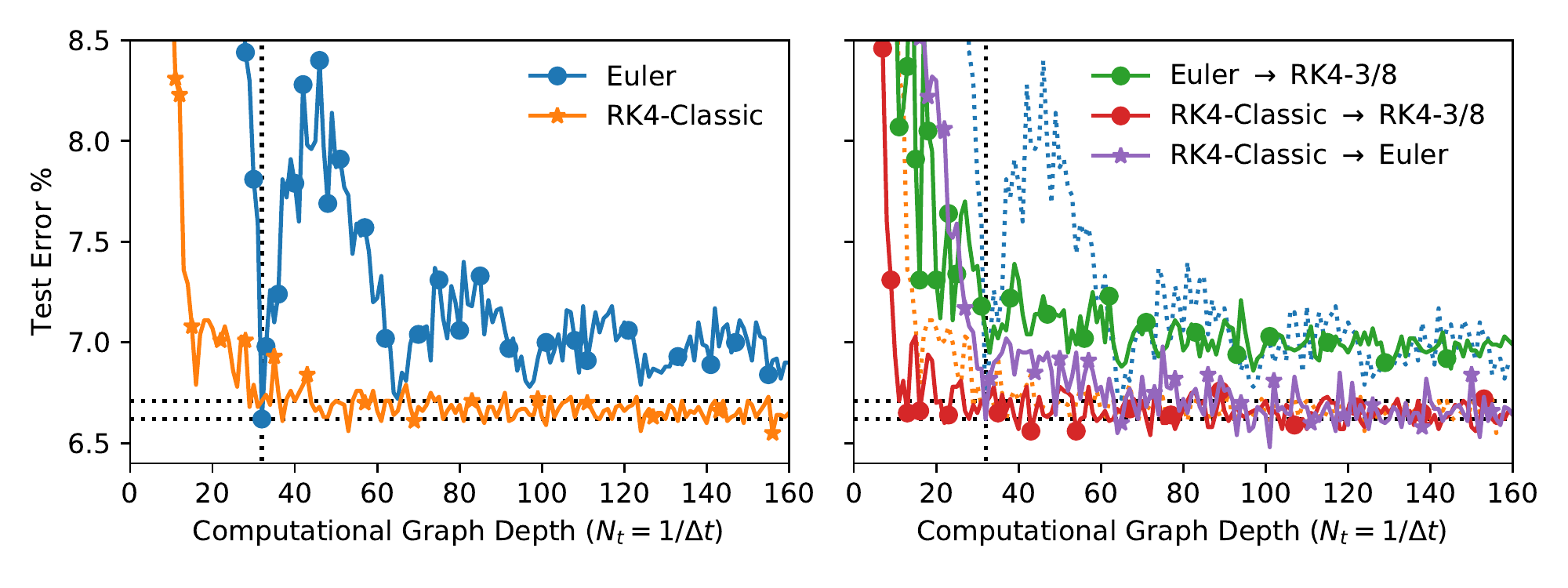}
		\end{overpic}\vspace{-0.4cm}	
		\caption{CIFAR-10, no skip initialization, Unit depth: 32-32-32, Channels per block: 16-32-64}
	\end{subfigure}\vspace{-0.0cm}
	
	\begin{subfigure}[t]{0.9\textwidth}
		\centering
		\begin{overpic}[width=1\textwidth]{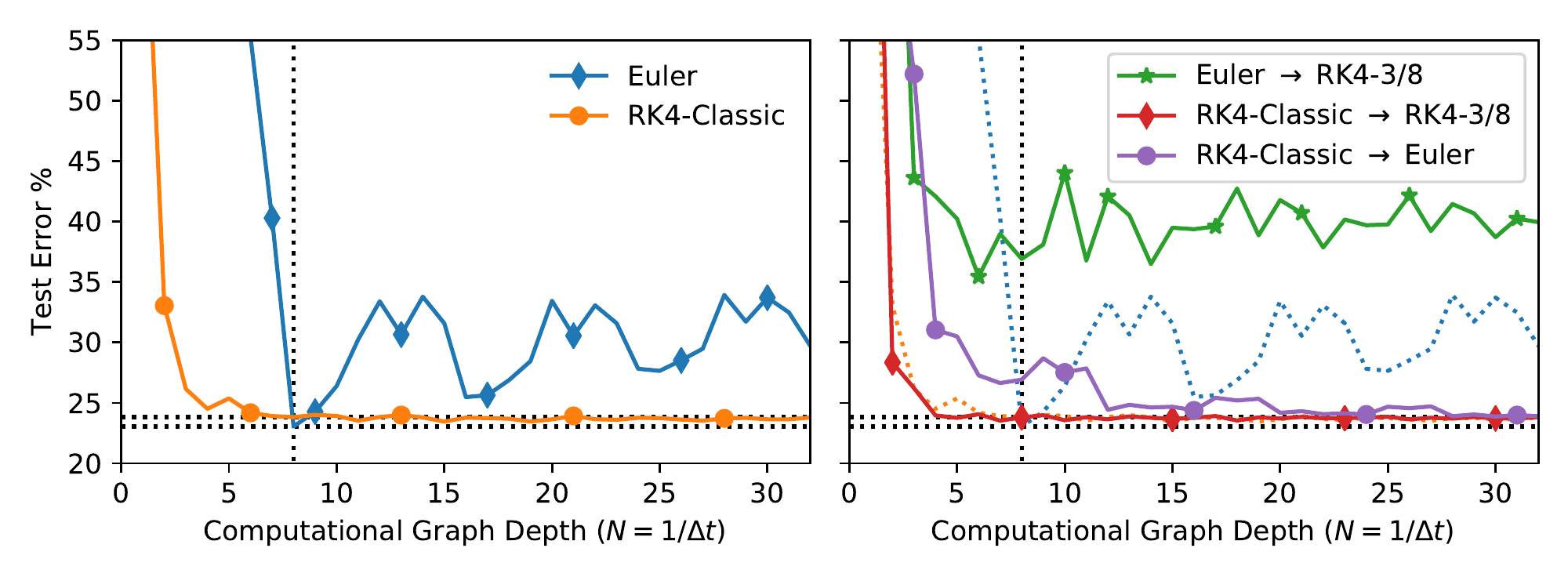} 
		\end{overpic}\vspace{-0.4cm}			
		\caption{CIFAR-100, no skip initialization, Unit depth: 8-8-8, Channels per block: 32-64-128}
	\end{subfigure}
	
	\begin{subfigure}[t]{0.9\textwidth}
		\centering
		\begin{overpic}[width=1\textwidth]{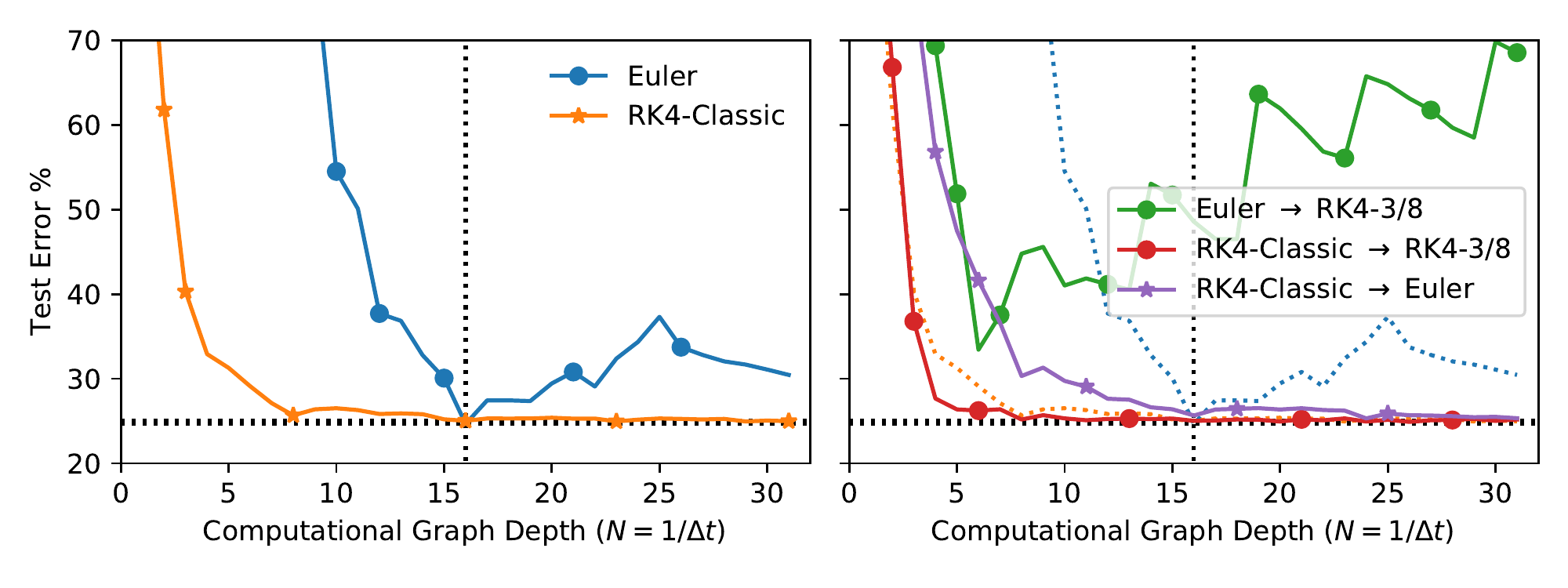} 
		\end{overpic}\vspace{-0.4cm}			
		\caption{CIFAR-100, no skip initialization, Unit depth: 16-16-16, Channels per block: 32-64-128}
	\end{subfigure}
	
	\caption{
		Additional variations of CIFAR-10 and CIFAR-100 that did not use the skip-initialization term. (a) and (b) are identical in all other respects to the models of Figure 4. In (c), the model is twice as deep but half as wide as the CIFAR-100 model discussed in the main text.  When batch-normalization is the main stabilization term in the network, the Euler model is more robust to alteration of time-step, but it still significantly decreases in accuracy. 
		\CoolNet{}(RK4-Classic) exhibits manifestation invariance in all cases with this architecture variant as well.
		%
	}
	\label{fig:cifar10_manifestation_appx}
\end{figure}

Here, we present results from applying \CoolNet{} to other variations of the image classification task, to extend the analysis of Section \ref{sxn:refinenet-empirical}.
We first apply the wide architecture to Tiny-Imagenet; and then we show three more configurations for CIFAR-10 and CIFAR-100, with variations on the residual module.
In each configuration, the Euler-based ResNet does not satisfy the properties of a dynamical system, but the higher-order trained \CoolNet{} exhibits manifestation~invariance.

\subsection{Tiny-ImageNet}

The \CoolNet{} model was also applied to the Tiny-ImageNet dataset, a version with $64\times 64$ pixel images and 200 classes. We did not fine-tune the architecture to this dataset: we used the same  architecture as for CIFAR-100, with $N_t=8$ and three OdeBlocks with 64, 128, and 256 features. 
The only difference is that the last fully-connected layer was modified to output 200 classes instead of 100. 
We compared the wide \CoolNet{}(Euler) with a wide \CoolNet{}(Midpoint).
See Fig.~\ref{fig:tiny} for results of the convergence test study.
Midpoint was chosen for practical run-time considerations, as the $64\times64$ require more computation cost than the CIFAR images.
The \CoolNet{}(Midpoint) was refined at epochs [20, 50, and 80] (the same schedule as for the CIFAR-100 experiment.)
The parameter $\epsilon$ was set to 1.
Because this architecture is designed to perform well on CIFAR, it only achieves $\approx 40\%$ performance on this more complex task.
As before, the Euler-based ResNet diverges after changing timestep size or integrator.
Training with the second order midpoint integrator is sufficient to exhibit the manifestation invariance property of \CoolNet{}.
This shows that our methodology also forces the \CoolNet{} to learn a dynamical system on larger image classification~problems.

\subsection{CIFAR-10 and CIFAR-100}

Three additional comparisons for CIFAR are shown in Fig.~\ref{fig:cifar10_manifestation_appx}.
In each of these cases, we do not use a skip-initialization term in the residual module, such that batch normalization is the only regularization operation.
This makes the final model less accurate.
In each case, the Euler model has a peak at the original timestep size but then lower accuracy for all variations, while \CoolNet{}(RK4-Classic) exhibits the same manifestation property illustrated in the main text.
An interesting observation is that, without the skip-initialization term, the divergence of the Euler model is alleviated, and it does not deviate towards 100\% error during time-step refinement, but it still looses significant accuracy.

